\documentclass[letterpaper]{article} 
\usepackage[preprint]{aaai2027}  
\usepackage[hyphens]{url}  
\usepackage{graphicx} 
\usepackage{natbib}  
\usepackage{caption} 
\usepackage{amsmath,amssymb,amsfonts}
\usepackage{subcaption}   
\usepackage{array}
\usepackage{booktabs}
\usepackage{multirow}
\usepackage{makecell}
\usepackage{tabularx}
\usepackage{threeparttable}

\providecommand{\href}[2]{#2}

\providecommand{\TODO}[1]{}

\title{HandAnthro: Automated Hand Anthropometry from a Single Image}

\author{
    Fan Zhou\textsuperscript{\rm 1},
    Shuairan Chen\textsuperscript{\rm 1},
    Mengying Zhang\textsuperscript{\rm 1},
    Yulin Wu\textsuperscript{\rm 1},\\
    Sadegh Jafari\textsuperscript{\rm 1},
    Sixing Yu\textsuperscript{\rm 2},
    Rui Li\textsuperscript{\rm 1},
    Ali Jannesari\textsuperscript{\rm 1},
    Guowen Song\textsuperscript{\rm 1}\corresponding
}
\affiliations{
    \textsuperscript{\rm 1}Iowa State University\\
    \textsuperscript{\rm 2}Microsoft\\
    \{fanzhou,shuairan,mezhang,yulinw,sadegh,ruili,jannesar,gwsong\}@iastate.edu\\
    sixingyu@microsoft.com
}

\begin{document}
\maketitle

\hyphenation{HandAnthro}
\begin{abstract}
Hand anthropometry supports protective-glove design, but existing measurement methods often require trained operators, specialized hardware, or manual landmarking. We present \textbf{HandAnthro}, which estimates 44 projected hand dimensions from a smartphone photograph of a palm-up hand on US letter-size paper. The pipeline reconstructs wrist-occluded paper boundaries for rectification, whitens non-hand pixels, and refines 41 anthropometry-specific landmarks from a fine-tuned You Only Look Once (YOLO) pose model using image-specific geometry and contours. Controlled evaluation comprised 720 captures from 45 held-out participants, each contributing 16 images across two smartphones, two backgrounds, two angles, and two nominal illumination settings. HandAnthro produced complete outputs for 704 captures (97.8\%); among these, mean absolute error (MAE) was $3.80$\,mm per dimension against two trained operators' caliper measurements. Regional MAEs were $2.48$\,mm for non-thumb fingers, $6.04$\,mm for thumbs, and $6.17$\,mm for palm and wrist. In a researcher-assisted mobile-app pilot, automated batch processing returned all 44 dimensions for 260 of 268 retained, researcher-screened firefighter images (97.0\%). A descriptive, unpaired comparison with an independent national firefighter reference yielded a mean absolute difference of $2.40$\,mm across 28 sex-by-dimension group-mean contrasts. These results characterize controlled measurement performance and researcher-assisted field feasibility for future distributed hand-anthropometry studies.
\end{abstract}

\section{Introduction}
Acute occupational hand injuries account for more than one million US emergency-department visits annually, and protective gloves reduce their relative risk by about 60\%~\cite{sorock2004case,sorock2004glove}. This protection depends on fit: poor fit compounds losses in dexterity, tactile sensitivity, and grip strength and reduces user acceptance~\cite{dianat2012gloves,griffin2018methods}. Yet glove-sizing systems often rely on outdated or unrepresentative hand data~\cite{hsiao2015firefighter_hand_Dim_Schema_design,hsiao2026firefighter}. Large, current anthropometric samples are therefore valuable for glove and hand-tool design, but direct caliper measurement requires trained operators, manual photogrammetry requires work on each image, and three-dimensional (3D) approaches require scanners or calibrated multi-view hardware~\cite{griffin2018dimensions,yu20132d,yang2021development}.

Prior automated two-dimensional (2D) hand-measurement systems have typically relied on tightly controlled acquisition setups, such as flatbed scanners or calibrated industrial-camera rigs~\cite{han2016automatic,nguyen2025analyzing}. HandAnthro instead uses a simple smartphone protocol in which a palm-up, finger-spread hand is photographed on a letter-size paper sheet. The workflow uses the sheet geometry to rectify the image and recover metric scale, then localizes the anatomical endpoints needed to compute 44 projected dimensions. By reducing on-site equipment requirements and enabling an automated image-to-measurement workflow without manual involvement, this design targets two major operational barriers to larger, geographically distributed occupational hand-anthropometry studies. First, the system must recover a reliable paper reference despite wrist occlusion and camera viewpoint variation. Second, it must accurately localize measurement endpoints on the true hand contour despite unpredictable shadow patterns around the hand boundary caused by variations in real-world illumination and capture conditions.

HandAnthro addresses the first problem by reconstructing the occluded paper mask before quadrilateral detection. It addresses the second by whitening the non-hand pixels, fine-tuning a You Only Look Once (YOLO) pose architecture~\cite{Jocher_Ultralytics_YOLO_2023,Jocher_Ultralytics_YOLO11_2024} to predict anthropometry-specific landmarks, and refining those predictions with image-specific finger geometry and contour constraints. These stages automatically produce the protocol-defined projected measurements without participant-specific calibration.

We evaluate HandAnthro in two complementary studies. A controlled repeated-capture study quantifies perspective-rectification and end-to-end completion, caliper-referenced measurement accuracy, and sensitivity across the tested capture conditions. A separate researcher-assisted mobile-app field pilot evaluates completion outside the capture rig and compares sex-specific group means with corresponding hand-anthropometry summaries from an independent national firefighter cohort. Because the cohorts are independent, this comparison assesses population-level plausibility rather than individual-level measurement accuracy.

\section{Related Work}
Classical hand anthropometry measures standardized dimensions directly with contact instruments such as calipers or rulers between predefined anatomical landmarks~\cite{gordon1989anthropometric,yu20132d}. The resulting staffing and quality-control burden makes distributed collection operationally demanding, although it does not preclude large centrally organized surveys.

Digital methods measure dimensions from 2D images or 3D surfaces. Controlled 2D studies compared software-derived hand dimensions with direct measurements, but used fixed camera geometry and interactive placement of calibration or measurement lines~\cite{habibi2013precise,patel2018validation_2DPhoto}. 2D and 3D scanning methods have likewise been compared with direct measurements for selected dimensions~\cite{li2008validation,yu20132d}. 3D acquisition commonly requires a scanner or calibrated multi-view system and additional landmarking~\cite{griffin2018methods,yang2021development}. Deep learning has also automated measurement extraction from a single 3D hand scan, although the input still depends on 3D sensing hardware~\cite{kaashki2022automatic}.

Prior automated 2D methods vary in acquisition requirements and measurement outputs. A freehand smartphone prototype estimated the length ratio between two fingers without scale~\cite{sandnes2014smartphone}. A reference-square workflow automatically estimated finger lengths from four digital-camera images per participant~\cite{magno2013input}. Systems reporting more dimensions have used more controlled acquisition: an enclosed flatbed-scanner system automatically derived 18 dimensions from one palmar scan~\cite{han2016automatic}, whereas a calibrated industrial-camera system derived 24 parameters from 12 images covering six postures of each hand in a light-controlled box~\cite{nguyen2025analyzing}. These systems demonstrate 2D automation but differ in output breadth, capture count, portability, environmental control, and use of dedicated rigs.

Table~\ref{tab:literature_measurement_context} summarizes automated dimension extraction, acquisition equipment, study samples, and source-reported mean absolute error (MAE) for hand dimensions. It includes methods that automate anatomical dimension extraction or report hand-dimension MAE against physical reference measurements.


\begin{table*}[t]
\centering
\footnotesize
\setlength{\tabcolsep}{3.5pt}
\renewcommand{\arraystretch}{1.15}
\begin{tabular}{@{}
  >{\raggedright\arraybackslash}p{0.11\textwidth}
  >{\raggedright\arraybackslash}p{0.24\textwidth}
  >{\raggedright\arraybackslash}p{0.14\textwidth}
  >{\raggedright\arraybackslash}p{0.19\textwidth}
  >{\raggedright\arraybackslash}p{0.25\textwidth}@{}}
\toprule
Method & Automation and equipment requirements & Study sample; dimensions & Reference & Hand-dimension MAE (mm) \\
\midrule
\citeauthor{han2016automatic} (\citeyear{han2016automatic}) &
\textbf{Automated}; required equipment: flatbed scanner and enclosure &
11 people; 17 tabulated dimensions &
Live-hand calipers &
--- \\
\addlinespace
\citeauthor{kaashki2022automatic} (\citeyear{kaashki2022automatic}) &
\textbf{Automated}; required equipment: depth sensor and tablet &
20 people; 11 real-hand dimensions &
Anthropometrist; instrument unspecified &
4.5 \\
\addlinespace
\citeauthor{nguyen2025analyzing} (\citeyear{nguyen2025analyzing}) &
\textbf{Automated}; required equipment: industrial camera, light box, ring light, and calibration target &
539 people; 24 dimensions &
Gauge-block calibration &
--- \\
\midrule
HandAnthro &
\textbf{Automated}; required equipment: smartphone and known-size paper &
45 people; 44 dimensions &
Mean of two operators' caliper readings &
3.80 overall;\newline
2.48 non-thumb fingers;\newline
6.04 thumb;\newline
6.17 palm/wrist \\
\bottomrule
\end{tabular}
\caption{Selected hand-anthropometry methods. MAEs are not directly comparable across studies; see Appendix~\ref{app:literature_comparison_notes} for definitions and reporting notes.}
\label{tab:literature_measurement_context}
\end{table*}

\section{System Design and Methodology}

\begin{figure}[!t]
\centering
\includegraphics[width=\columnwidth]{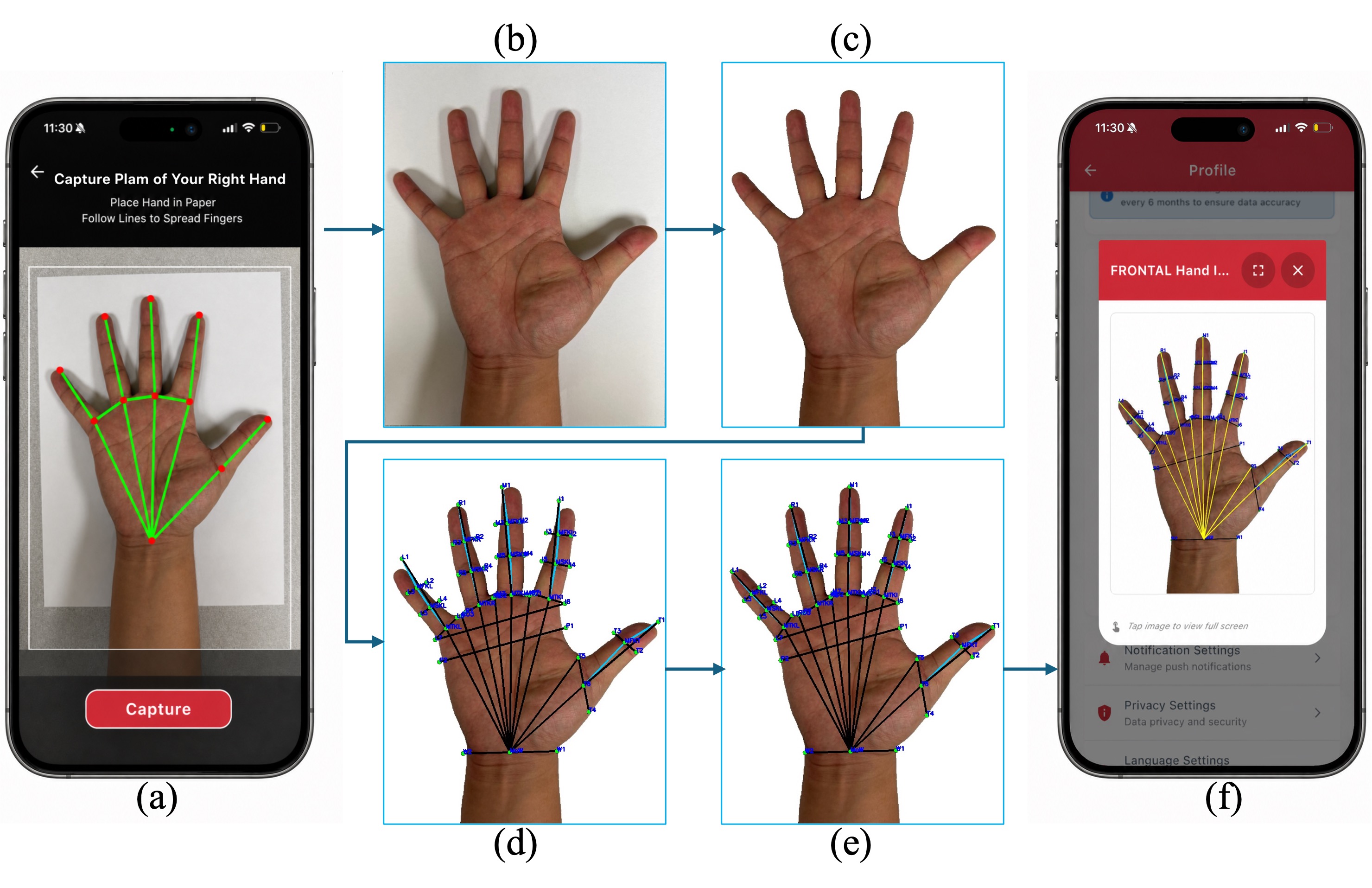}
\caption{HandAnthro capture-to-measurement workflow.}
\label{Fig_1.Pipeline_overview}
\end{figure}

\noindent\textbf{Capture-to-Measurement Workflow.}
Figure~\ref{Fig_1.Pipeline_overview} presents the capture-to-measurement workflow. Panel~(a) shows the aisafehand mobile app used to acquire the required hand-on-paper image. Perspective rectification (PR) uses the recovered paper quadrilateral to produce a fronto-parallel view, correcting projective distortion and providing the geometric reference for metric scaling~(b). Background whitening (BW) sets non-hand pixels to white, reducing interference from variable shadows around the hand and between the fingers~(c). A YOLO landmark predictor produces 41 initial anthropometry-specific landmarks~(d); geometry-constrained post-processing (PP) refines them using finger geometry and contour evidence~(e); and the app visualizes the resulting 44 predicted dimensions~(f).

\begin{figure}[!htbp]
\centering
\includegraphics[width=\columnwidth]{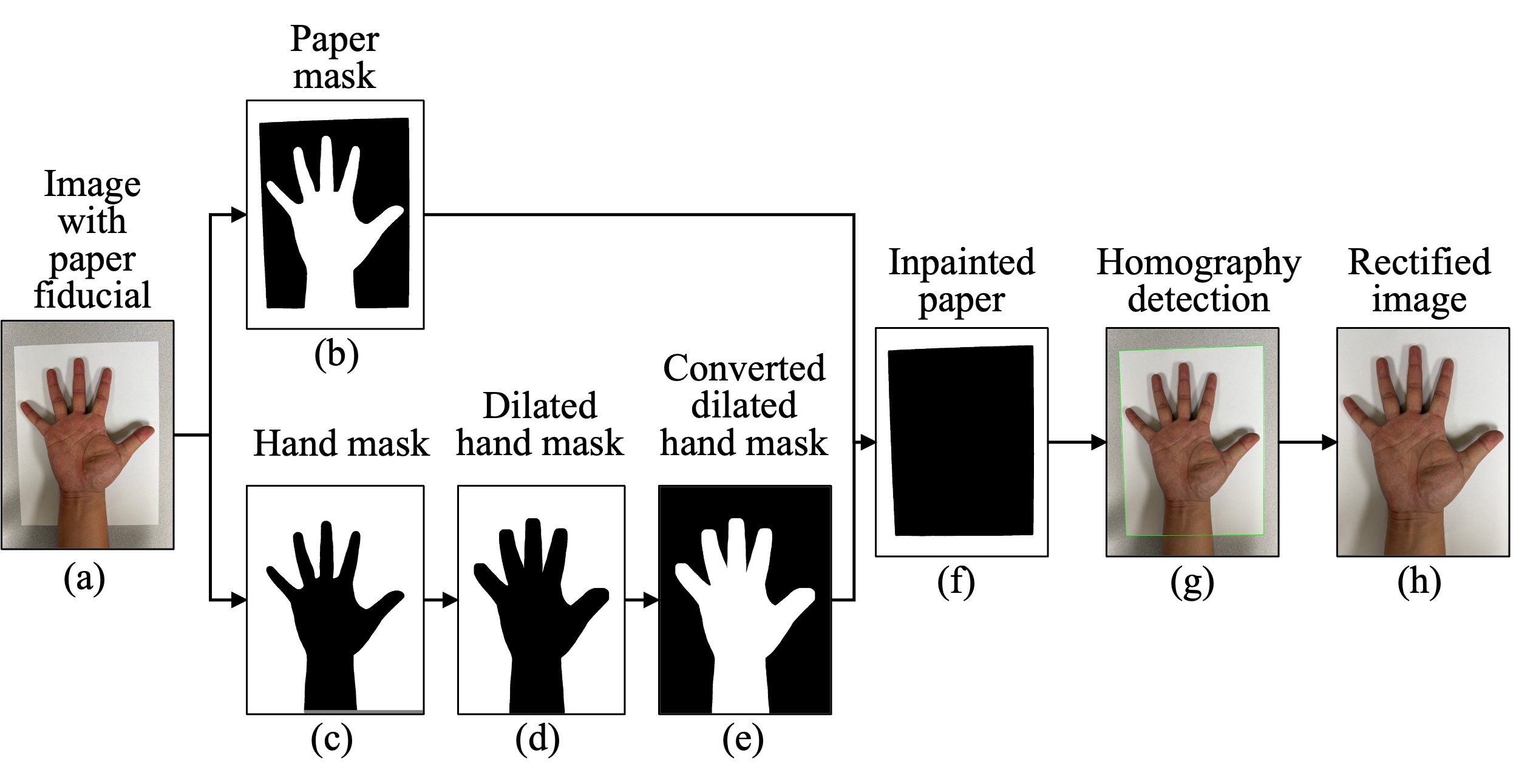}
\caption{Occlusion-aware PR workflow: (a)~input image; (b,c)~initial paper and hand masks; (d,e)~construction of the inpainting region; (f)~completed paper mask; (g)~detected paper quadrilateral (green frame); and (h)~perspective-rectified RGB image.}
\label{fig:pr_workflow_main}
\end{figure}

\noindent\textbf{Occlusion-Aware Perspective Rectification.}
\label{sec:Perspective Rectification}
The paper reference defines the target plane for correcting projective distortion and supplies the known geometry used for metric scaling. Because the wrist interrupts a paper boundary, HandAnthro uses the reconstruct-then-detect sequence illustrated in Fig.~\ref{fig:pr_workflow_main}. Given an input image (Fig.~\ref{fig:pr_workflow_main}~(a)), Segment Anything in High Quality (SAM-HQ)~\cite{ke2023segment} segments an incomplete paper mask (Fig.~\ref{fig:pr_workflow_main}~(b)) and a raw hand mask (Fig.~\ref{fig:pr_workflow_main}~(c)).

Fixed prompt coordinates for mask segmentation are unsuitable because the location and scale of the hand within the image, as well as its placement on the paper, vary across captures. In the PR stage, HandAnthro derives image-adaptive SAM-HQ prompts from the 21 hand-pose landmarks returned by MediaPipe Hands~\cite{zhang2020mediapipe}. These auxiliary hand-pose landmarks provide coarse hand-location and scale cues for prompt placement. Appendix~\ref{app:samhq_prompts} documents the evaluated adaptive prompt construction.

The raw hand mask is dilated once with a \(31\times31\) square kernel so that the inpainting region fully covers the portion of the paper occluded by the hand and wrist (Fig.~\ref{fig:pr_workflow_main}~(d)). The dilated mask is then polarity-converted so that its white foreground identifies the region supplied to LaMa~\cite{suvorov2022resolution} for inpainting (Fig.~\ref{fig:pr_workflow_main}~(e)). LaMa completes the hand-shaped interruption in the paper-mask representation, producing a recovered paper mask (Fig.~\ref{fig:pr_workflow_main}~(f)).

The completed paper mask (Fig.~\ref{fig:pr_workflow_main}~(f)) is used to detect the paper quadrilateral and its four corners. The detected quadrilateral is overlaid on the original red-green-blue (RGB) image for visualization (Fig.~\ref{fig:pr_workflow_main}~(g)). Matching the four detected source corners to the corresponding corners of an ideal rectangle with the known paper aspect ratio determines the homography \(\mathbf H\). Applying \(\mathbf H\) to the original RGB image produces the perspective-rectified image (Fig.~\ref{fig:pr_workflow_main}~(h)).

\noindent\textbf{Background Whitening and Anthropometry-Specific Landmarks.}
HandAnthro reuses the raw, undilated hand mask produced during PR (Fig.~\ref{fig:pr_workflow_main}~(c)). The same homography \(\mathbf H\) warps the original RGB image and the raw hand mask into the rectified frame, preserving their pixel correspondence. In the rectified image (Fig.~\ref{fig:pr_workflow_main}~(h)), BW retains RGB values inside the warped hand mask and sets all pixels outside it to white. This removes non-hand pixels, including shadows, strengthening the contour evidence without a second segmentation inference (Appendix~\ref{app:bw_masks}). The background-whitened image is resized to a canonical \(720\times932\)-pixel frame and passed to the YOLO predictor.

We define and annotate 41 anthropometry-specific landmarks spanning the thumb, four fingers, inter-finger roots, palm, and wrist (Appendix~\ref{app:keypoints}). The final predictor is a YOLOv11x-pose model~\cite{Jocher_Ultralytics_YOLO11_2024} fine-tuned with augmentation on background-whitened images using this landmark schema; Section~\ref{sec:exp_setup} describes the evaluated model configurations.

\noindent\textbf{Geometry-Constrained Refinement.}
\label{post processing}
PP refines the anatomical placement of measurement endpoints. Initial YOLO finger landmarks can misalign with the finger's orientation and length. PP uses MediaPipe hand-pose landmarks and the hand contour from the background-whitened image to estimate each finger's axis. A forward search from the finger centerline along this axis locates a corrected fingertip on the contour.

\begin{figure}[!t]
\centering
\includegraphics[width=0.7\columnwidth]{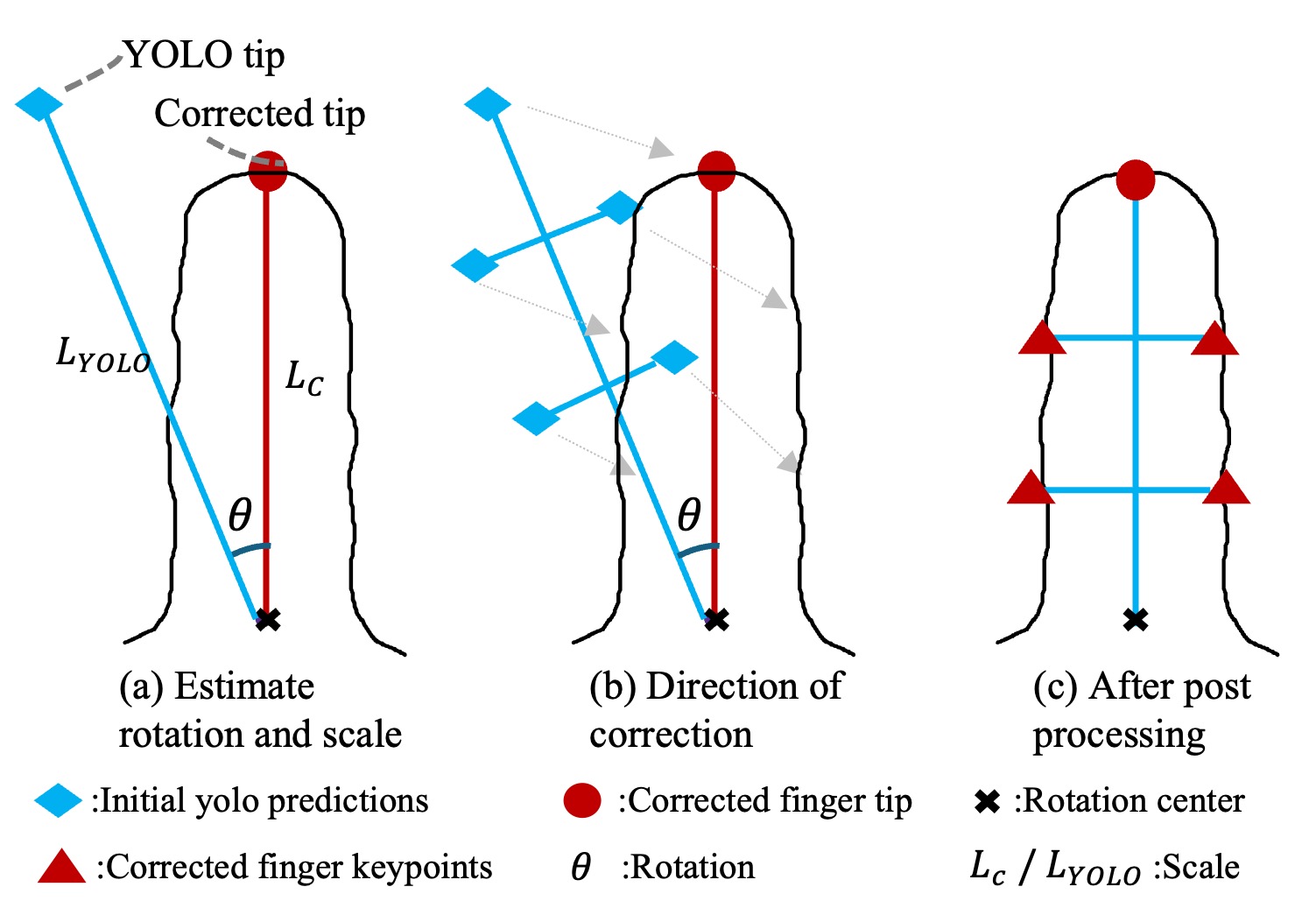}
\caption{Geometry-constrained correction of groupwise rotation-and-scale misalignment.}
\label{fig:rotation_scale_correction}
\end{figure}

For each finger, PP holds the YOLO-predicted root fixed and compares the predicted and corrected root-to-tip vectors (Fig.~\ref{fig:rotation_scale_correction}). Their angular difference defines the rotation \(\theta\), and their length ratio \(L_c/L_{\mathrm{YOLO}}\) defines the scale factor. PP applies this rotation and scaling to the finger's configured landmark group, then aligns eligible lateral endpoints with the hand boundary along their measurement directions (Appendix~\ref{app:pp_formulation}).

\noindent\textbf{Metric Conversion.}
\label{sec:scale_estimation}
After PP, 44 fixed endpoint pairs define the reported dimensions.
During PR, the four detected paper corners are mapped to a
fronto-parallel rectangle with the US Letter width-to-height ratio
of \(8.5:11\). The rectified image is resampled to the canonical
\(720\times932\)-pixel frame, so the sheet's 11-inch height corresponds
to 932 pixels and defines a scale of \(932/11\) pixels per inch.
For dimension \(d\), \(\widehat y_d=\frac{25.4}{932/11}\lVert u_d-v_d\rVert_2\)\,mm, where \(u_d\) and \(v_d\) are its final endpoints.

\section{Experiments and Evaluation}
\label{sec:exp_setup}
\noindent\textbf{YOLO Fine-Tuning and Ablation Design.}
The model-development dataset comprised one annotated palmar image from each of 42 independent participants. The participants were randomly assigned to mutually exclusive training and validation sets comprising 33 and 9 images, respectively. The 45 participants in the controlled test set were held out from both sets, with no participant overlap. To isolate the incremental contributions of PR, BW, and PP, we evaluated four progressively constructed pipeline configurations. In ablation case \textbf{A0}, YOLO was fine-tuned on the original captured images. In \textbf{A1}, YOLO was fine-tuned on perspective-rectified images. In \textbf{A2}, YOLO was fine-tuned on images processed by both PR and BW. \textbf{A3} then applied PP to the corresponding A2 predictions. Each configuration was evaluated using two pose backbones (YOLOv8x-pose~\cite{Jocher_Ultralytics_YOLO_2023} and YOLOv11x-pose) and two training data augmentation settings (with and without augmentation), yielding \(4 \times 2 \times 2 = 16\) experiments (see Appendices~\ref{app:ablation_deltas} and~\ref{app:hparams} for details).

\noindent\textbf{Controlled Repeated-Capture Evaluation.}
The controlled test set comprised 45 held-out participants, each contributing 16 palmar captures spanning all combinations of two smartphones, two backgrounds, two capture angles, and two nominal illumination settings (720 images). We evaluated PR and end-to-end completion over all 720 inputs. The two nominal illumination settings produced little visible difference in brightness in the captured images, as the smartphones' built-in automatic exposure and image processing adjusted image brightness across conditions. We therefore pooled the two settings and did not assess illumination effects separately. Capture examples and rig geometry are shown in Appendices~\ref{app:capture_conditions} and~\ref{app:capture_rig}.

Two trained operators independently measured all 44 dimensions of each participant's hand using calipers; their mean served as the reference for each dimension (Appendix~\ref{app:manual_reliability}). Automated measurement accuracy was evaluated on the 704 complete outputs. We first calculated each dimension's MAE across these images, then averaged the 44 dimension-specific MAEs to obtain overall MAE. Regional MAEs used the same aggregation within the thumb (D1--D5), four non-thumb fingers (D6--D33), and palm and wrist (D34--D44).

For capture-condition accuracy, we calculated each dimension's MAE separately for each smartphone, background, and angle setting, then averaged across the 44 dimensions (Appendix~\ref{app:robustness_table}). For stability, we calculated the standard deviation (SD) of each participant's predictions for each dimension across completed captures. For each participant and dimension, we also calculated the absolute difference between the mean predictions at the two levels of each factor, then averaged these differences across participants and dimensions (Appendix~\ref{app:within_person}).

For landmark-based evaluation, we selected one image per participant from the 704 complete captures, aiming for balanced representation across the eight smartphone--background--angle combinations. The 41 landmarks were manually annotated on each selected image after PR and BW. We report the mean Euclidean pixel error across the \(45\times41\) predicted--reference landmark pairs. From these same annotations, we derived all 44 dimensions, then compared them with the same caliper reference. This annotation-to-caliper diagnostic comprised 1,980 image--dimension comparisons and was reported separately from the 704-image automated evaluation (Appendix~\ref{app:oracle_analysis}).

\noindent\textbf{Stage-Level Baseline Comparisons.}
We restricted stage-level baselines to alternatives that produced directly comparable outputs at the evaluated pipeline stage. Canny--Hough perspective rectification (CH-PR), our baseline using Canny edge detection and probabilistic Hough line detection~\cite{canny1986computational,matas2000robust}, replaced HandAnthro's complete PR module and was evaluated for completion over all 720 controlled captures; its measurement-error comparison used the 406 captures completed by both pipelines (Appendix~\ref{app:baseline_prcv}). For the BW baselines, we separately substituted REMBG~\cite{Gatis_rembg_2025}, BackgroundRemover~\cite{nader_backgroundremover_pypi}, or CarveKit~\cite{selin2024carvekit_github} to process the 704 captures for which HandAnthro completed PR, while holding PR, landmark prediction, and PP fixed. End-to-end measurement error for each BW variant was calculated on that variant's complete outputs; BW-stage runtime was also measured. Section~\ref{subsec:ablation_results} reports these comparisons.

\noindent\textbf{Researcher-Assisted Field Pilot.}
At multiple professional events for firefighters, staff assisted with app operation, hand placement, and photography for the field pilot. The cohort comprised 268 firefighters (204 male, 64 female), each with one retained dominant-hand palmar image. Unsuitable photographs were discarded and retaken. Retained images underwent automated batch processing on a laboratory desktop without human intervention. All completed outputs were included in the analysis without human correction (Appendix~\ref{app:hsiao}).

Processing completion was defined as returning all 44 dimensions and was evaluated over the 268 retained images. Sex-specific means for 14 dimensions were compared with published summaries from an independent national firefighter cohort~\cite{hsiao2015firefighter_hand_Dim_Schema_design}, yielding 28 contrasts. This comparison assessed population-level plausibility.

\section{Results}
\subsection{Component Selection, Ablations, and Stage-Level Baselines}
\label{subsec:ablation_results}
\label{baseline}
\label{subsec:pixel_validation}
Figure~\ref{fig:supp_ablation_outputs} in Appendix~\ref{app:pixel_breakdown} shows representative outputs for ablation configurations A0--A3.

\begin{figure}[!t]
    \centering
    \subcaptionbox{HandAnthro\label{vis: OurPR}}{
        \includegraphics[height=0.28\columnwidth]{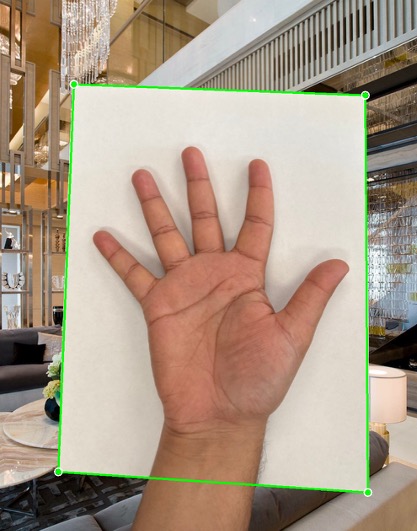}
    }
    \hfil
    \subcaptionbox{CH-PR\label{vis: BasePR}}{
        \includegraphics[height=0.28\columnwidth]{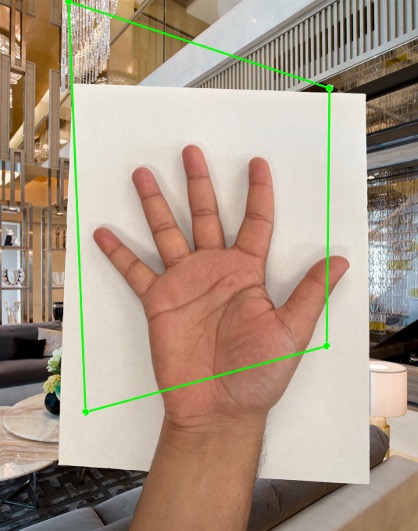}
    }
    \hfil
    \subcaptionbox{MAE Results\label{vis: PRMAE}}{
        \includegraphics[height=0.28\columnwidth]{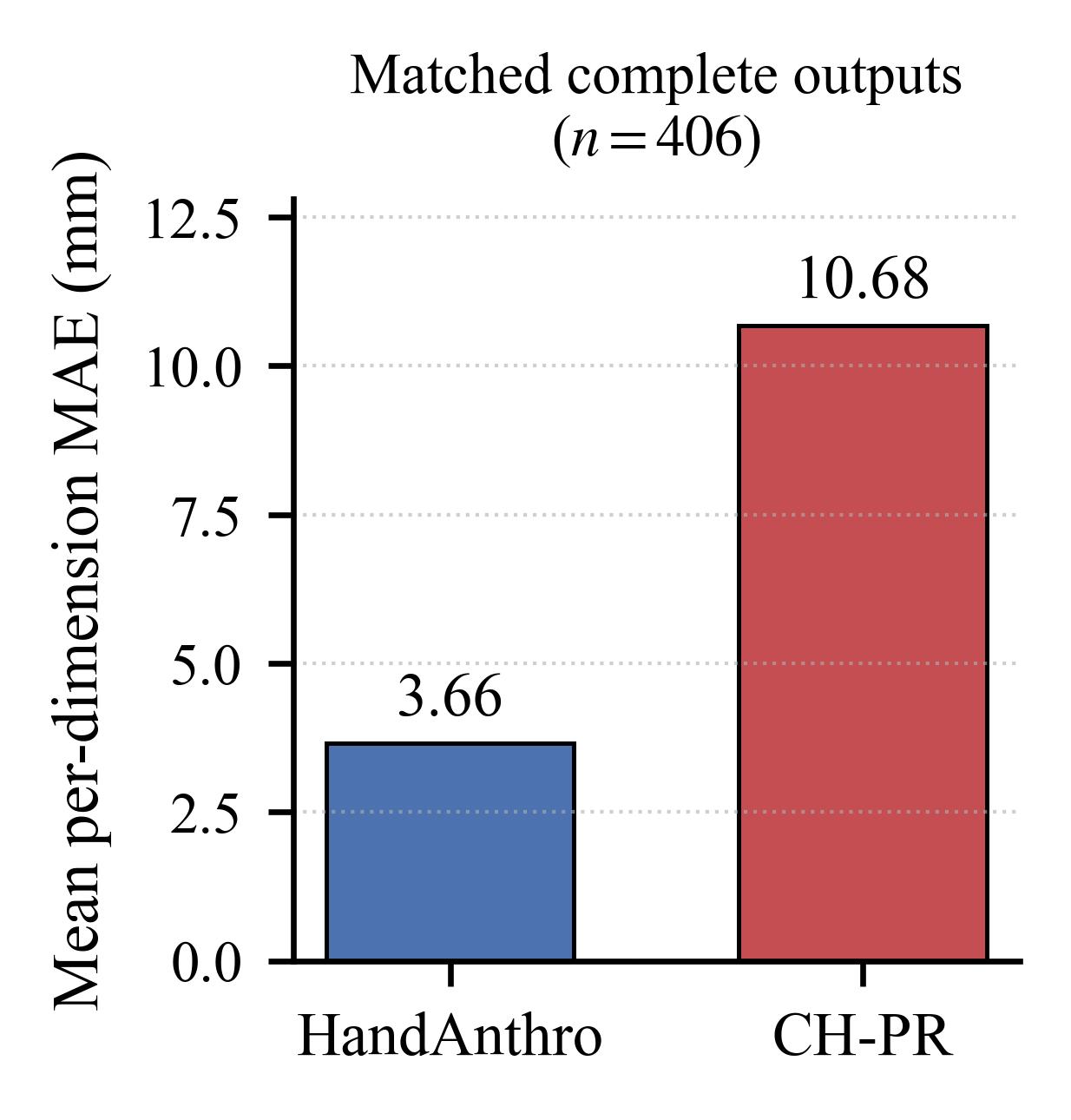}
    }
    \caption{Perspective-rectification comparison.}
    \label{vis: pr compare}
\end{figure}

\noindent\textbf{Perspective rectification.}
Across four matched backbone--augmentation configurations, A1 reduced YOLO validation pose loss~\cite{Jocher_Ultralytics_YOLO_2023,Jocher_Ultralytics_YOLO11_2024} by 77--94\% relative to A0. HandAnthro also completed PR on 704/720 captures, compared with 412/720 for CH-PR. On the 406 captures completed by both pipelines, MAE was 3.66\,mm for HandAnthro and 10.68\,mm for CH-PR (Fig.~\ref{vis: pr compare}).

\noindent\textbf{Background whitening.}
Adding BW after PR (A1 to A2) slightly increased validation pose loss for YOLOv8 and decreased it for YOLOv11 under both augmentation settings (Appendix~\ref{app:ablation_deltas}). We compared HandAnthro's BW, which reuses the hand mask from PR, with REMBG, BackgroundRemover, and CarveKit, leaving PR, the landmark predictor, and PP unchanged. All four methods received the same 704 rectified images, and each pipeline returned all 44 dimensions for every image. Among these 704 images, MAE against caliper measurements was 3.80\,mm with HandAnthro's BW and 4.26--4.71\,mm with the alternatives. Mean processing time for the BW stage alone was 17\,ms per image for HandAnthro and 93--379\,ms for the alternatives (Appendix~\ref{app:baseline_bg}).

\noindent\textbf{Landmark-model selection.}
Among the 12 A0--A2 training configurations, the augmented YOLOv11x-pose model had the lowest A2 validation pose loss, which was 1.5\% above the overall A1 minimum. We selected it because A3 requires the background-whitened contour for PP; full configuration results appear in Appendix~\ref{app:ablation_deltas}.

\noindent\textbf{Geometry-constrained refinement.}
On 45 annotated test images, PP reduced mean 2D landmark error
from 11.18 to 7.83\,px (29.9\% improvement); mean error decreased
on 44 of the 45 images. On the 704 captures completed by both
configurations (45 participants), caliper-referenced MAE
increased slightly from 3.707\,mm without PP to 3.804\,mm with PP.
The caliper reference is also subject to measurement uncertainty,
and dimension MAE alone does not fully characterize anatomical
landmark placement: different endpoint locations can produce
similar distances. Despite this 0.097\,mm (2.62\%) increase in dimension
MAE, we retain PP in the final framework to improve anatomical
landmark placement and support users' visual inspection (Fig.~\ref{Fig_1.Pipeline_overview}~(f)).

\subsection{Controlled Repeated-Capture Evaluation}
\label{subsec:controlled_results}

\noindent\textbf{Pipeline completion.}
HandAnthro completed PR and all 44 measurements for 704/720 controlled captures (97.8\%); no post-PR failures occurred.

\noindent\textbf{Measurement accuracy conditional on completion.}
On the 704 complete outputs, the mean of the 44 caliper-referenced MAEs was 3.80\,mm. Anatomical stratification yielded 2.48\,mm for the 28 non-thumb finger dimensions, 6.04\,mm for the five thumb dimensions, and 6.17\,mm for the 11 palm-and-wrist dimensions (details in Appendix~\ref{app:dim_breakdown}).

\noindent\textbf{Annotation-derived agreement with calipers.}
Dimensions computed from the manual image annotations had a caliper-referenced MAE of 3.38\,mm across 1,980 hand--dimension pairs (45 images and 44 dimensions; Appendix~\ref{app:oracle_analysis}).

\begin{figure}[!t]
\centering
\includegraphics[width=\columnwidth]{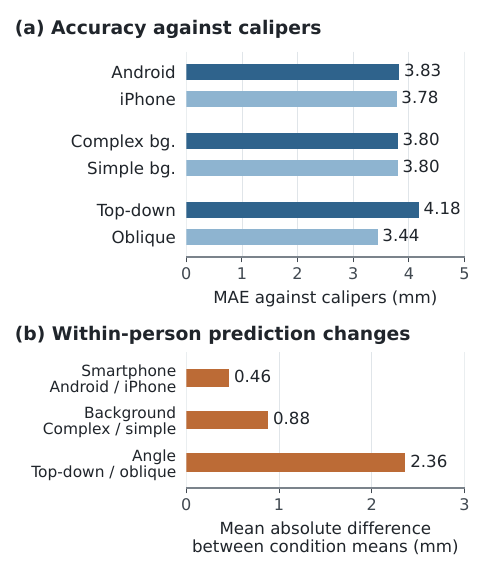}
\caption{Capture-condition accuracy and stability in 704 completed captures from 45 participants. (a) Caliper-referenced MAE averaged across 44 dimensions for each condition. (b) Absolute differences between the two condition-specific mean predictions, computed for each participant and dimension and then averaged across participants and dimensions. Both panels summarize completed outputs; seven participants had incomplete capture-condition grids.}
\label{fig:capture_condition_accuracy_stability}
\end{figure}

\paragraph{Measurement accuracy and stability across capture conditions.}
\label{subsec:robustness}
Among completed captures, aggregate MAEs were similar between the two tested smartphones and between background settings; top-down captures had higher MAE than oblique captures (Fig.~\ref{fig:capture_condition_accuracy_stability}~(a)).

Across 1,980 participant--dimension combinations (45 participants, 44 dimensions), within-person SD had a median of 1.25\,mm and a 95th percentile of 3.99\,mm. Mean absolute differences between each participant's condition-specific mean predictions were largest for angle and smallest for smartphone (Fig.~\ref{fig:capture_condition_accuracy_stability}~(b)). Per-dimension summaries and caliper-referenced angle contrasts appear in Appendices~\ref{app:within_person} and~\ref{app:angle_breakdown}.

\subsection{Researcher-Assisted Mobile-App Field Pilot}
\label{subsec:hsiao}
HandAnthro returned all 44 dimensions for 260/268 retained images (97.0\%). Among the complete cases (197 male and 63 female), comparison with an independent national firefighter reference yielded a mean absolute difference of 2.40\,mm across 28 sex-by-dimension group-mean contrasts~\cite{hsiao2015firefighter_hand_Dim_Schema_design} (Appendix~\ref{app:hsiao}).

\subsection{Failure Analysis}
\label{subsec:failure_analysis}
All failures occurred during PR (16/720 controlled; 8/268 field), and every PR-complete capture yielded all 44 measurements. Post-hoc review associated every controlled failure with wrist placement at or beyond a paper edge. Five field failures involved hands placed too high or low, with adaptive paper prompts outside the sheet. The remaining field failures involved incomplete paper framing, a hand extending beyond the sheet, or a covered corner (Appendix~\ref{app:failure_modes}).

\section{Discussion}
\label{sec:deployment}
\paragraph{Principal Findings.}
HandAnthro achieved 97.8\% controlled completion and 3.80\,mm
caliper-referenced MAE. PR standardized input geometry for
landmark prediction, and its addition reduced validation pose
loss by 77--94\% across matched configurations. BW supplied the
hand contour required by PP. Although its effect on validation
pose loss varied by backbone, the implementation that reused
the PR-stage mask yielded lower caliper-referenced MAE and
shorter runtime than the three tested BW alternatives.
PP reduced mean 2D landmark localization error by 29.9\%, while increasing
caliper-referenced MAE by 2.62\%. We retain PP to improve
anatomical landmark placement and support users' visual
inspection of where each measurement is taken.

\paragraph{Interpreting Measurement Agreement.}
The overall MAE of 3.80\,mm combines dimensions with different
error levels. MAE was 2.48\,mm for the 28 non-thumb finger
dimensions, 6.04\,mm for the five thumb dimensions, and
6.17\,mm for the 11 palm-and-wrist dimensions. The overall average
therefore does not represent the same level of agreement
across all anatomical regions.

For comparison, \citet{kaashki2022automatic} reported a mean
MAE of 4.5\,mm across 11 dimensions from 20 real hand scans.
HandAnthro's 3.80\,mm result is on the same broad scale of
several millimeters as this published result. However,
differences in dimension definitions, study samples, and
reference measurements prevent a direct accuracy ranking.
Table~\ref{tab:literature_measurement_context} provides this
measurement context alongside the methods' automation and
equipment requirements.

Dimensions derived from manually annotated landmarks still
had an MAE of 3.38\,mm against the caliper reference.
Thus, using manual annotations did not eliminate discrepancies
between image-derived dimensions and caliper measurements.
Possible contributors include endpoint-definition differences,
image projection, and uncertainty in annotation, rectification,
and caliper measurement.

\paragraph{Practical Implications.}
The capture-condition analyses describe both agreement with calipers and the stability of predictions for the same person. Across the tested settings, average within-person prediction changes were larger for angle than for smartphone or background (Fig.~\ref{fig:capture_condition_accuracy_stability}~(b)). This pattern motivates clearer capture-angle guidance and evaluation of whether such guidance improves measurement consistency.

The trained operator estimated that it would take about 15 minutes to measure all 44 dimensions of one hand with calipers. Once a photograph was available, HandAnthro extracted 44 projected dimensions with mean processing times of about 4 seconds on an NVIDIA RTX 4000 Ada graphics processing unit (GPU) and 26 seconds on an Apple M3 processor. These processing times exclude image collection. A smartphone and letter-size paper suffice for image capture, while automated extraction removes the need for manual landmark placement or for a trained operator to measure each dimension with calipers. Together, the simple capture requirements and short processing times offer potential for collecting hand anthropometric data from larger populations.

\paragraph{Limitations.}
Evaluation is limited to 45 controlled participants and a researcher-assisted field cohort without matched caliper measurements. Generalization to unassisted capture, additional devices, sites, and occupational domains remains to be evaluated.

\paragraph{Path to Deployment.}
Building on these results, we will evaluate unassisted app use, retaining all captures to assess attempts per user, guidance adherence, and completion rates overall and among compliant images. Planned AWS GPU inference will return measurements or recapture feedback, with response time and prediction quality evaluated. Pilot participants' concerns about fingerprint exposure in submitted palmar images motivate future evaluation of dorsal images as an alternative input for hand anthropometry. To address dimension-dependent biases, particularly in palm and wrist measurements, we will investigate offset and scale corrections using paired image-derived and caliper measurements, comparing shared and occupation-specific calibration on independent participants.
\section*{Ethics Statement and Resource Availability}
The institutional review board (IRB) deemed this study exempt under 45\,CFR\,46.104(d)(3)(i)(B). Potentially identifying participant images and individual-level data are not publicly released. The supplementary material provides the code, runnable sample, schemas, and aggregate results. Access to individual-level data requires prior IRB modification review; qualified researchers may contact the authors.

\section*{Acknowledgments}
This work was supported by the Fire Prevention and Safety (FP\&S) Research and Development Grant Program, administered by the Federal Emergency Management Agency (FEMA), U.S. Department of Homeland Security, under Award No.~EMW-2022-FP-00162. 

This work used the Delta system at the National Center for
Supercomputing Applications through allocation CIS240389 from the
Advanced Cyberinfrastructure Coordination Ecosystem: Services \& Support
(ACCESS) program, which is supported by National Science Foundation
grants \#2138259, \#2138286, \#2138307, \#2137603, and
\#2138296~\cite{boerner2023access}. This research used the Delta advanced computing and data resource which is supported by the National Science Foundation (award OAC 2005572) and the State of Illinois. Delta is a joint effort of the University
of Illinois Urbana-Champaign and its National Center for
Supercomputing Applications.

\nocite{douglas1973algorithms}
\bibliography{references}


\end{document}


\maketitle

\appendix

\hyphenation{HandAnthro}
\section{Test-Set Capture Conditions}
\label{app:capture_conditions}
\begin{figure*}[!tp]
\centering
\subcaptionbox{iPhone, simple background, top-down\label{cond:a}}[0.23\textwidth]{\includegraphics[width=0.18\textwidth]{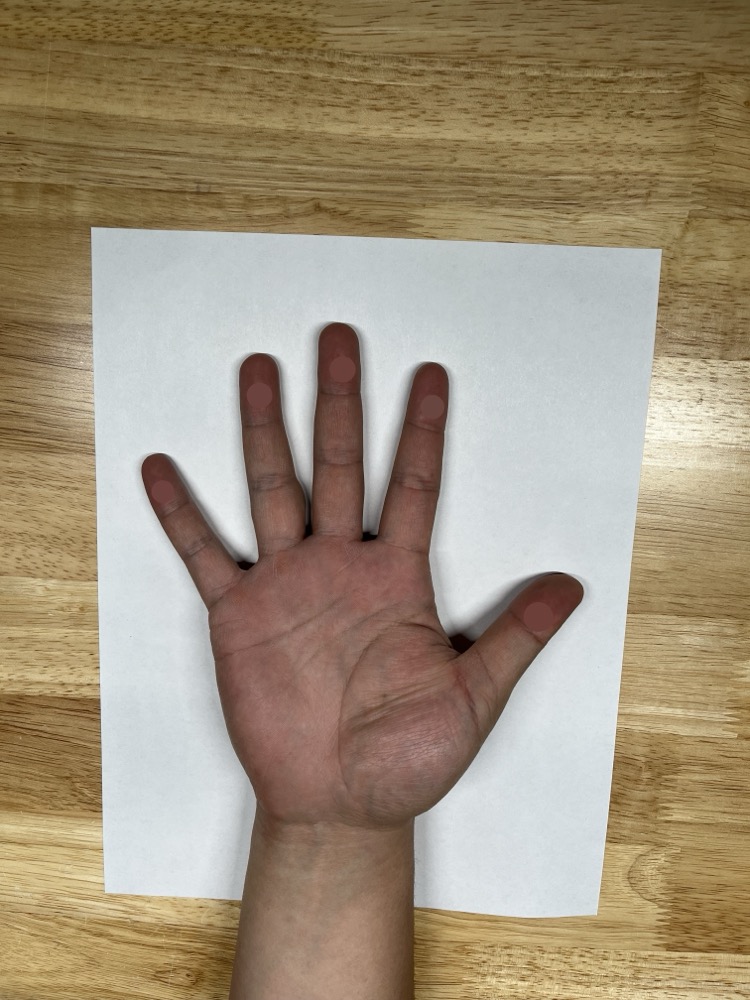}}\hfil
\subcaptionbox{iPhone, complex background, top-down\label{cond:b}}[0.23\textwidth]{\includegraphics[width=0.18\textwidth]{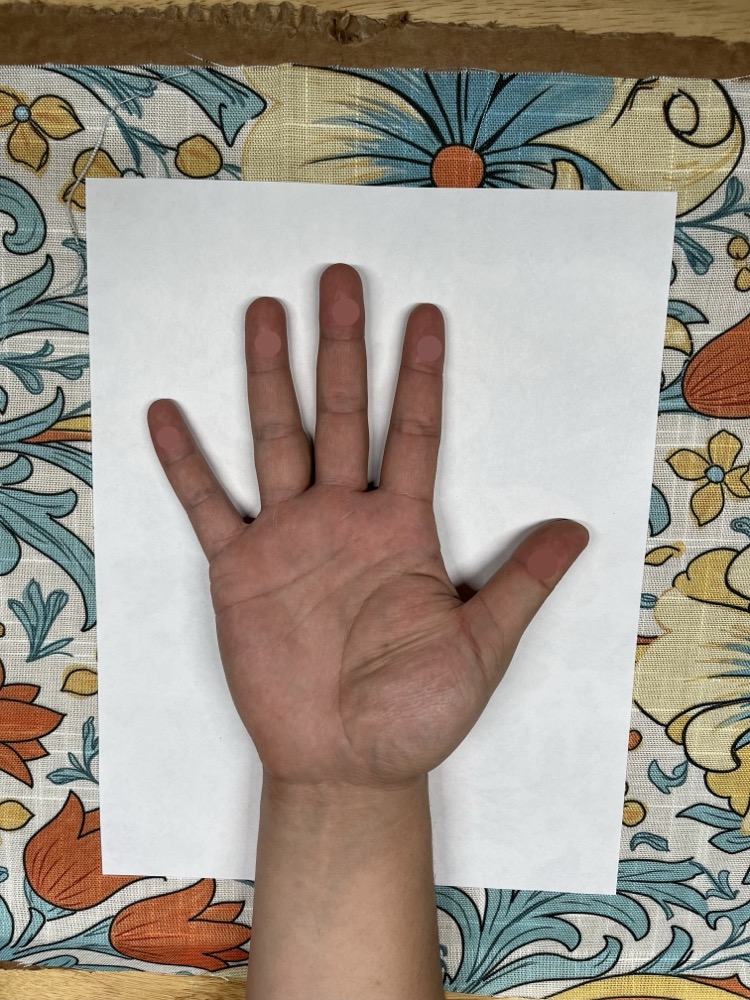}}\hfil
\subcaptionbox{Android, simple background, top-down\label{cond:c}}[0.23\textwidth]{\includegraphics[width=0.18\textwidth]{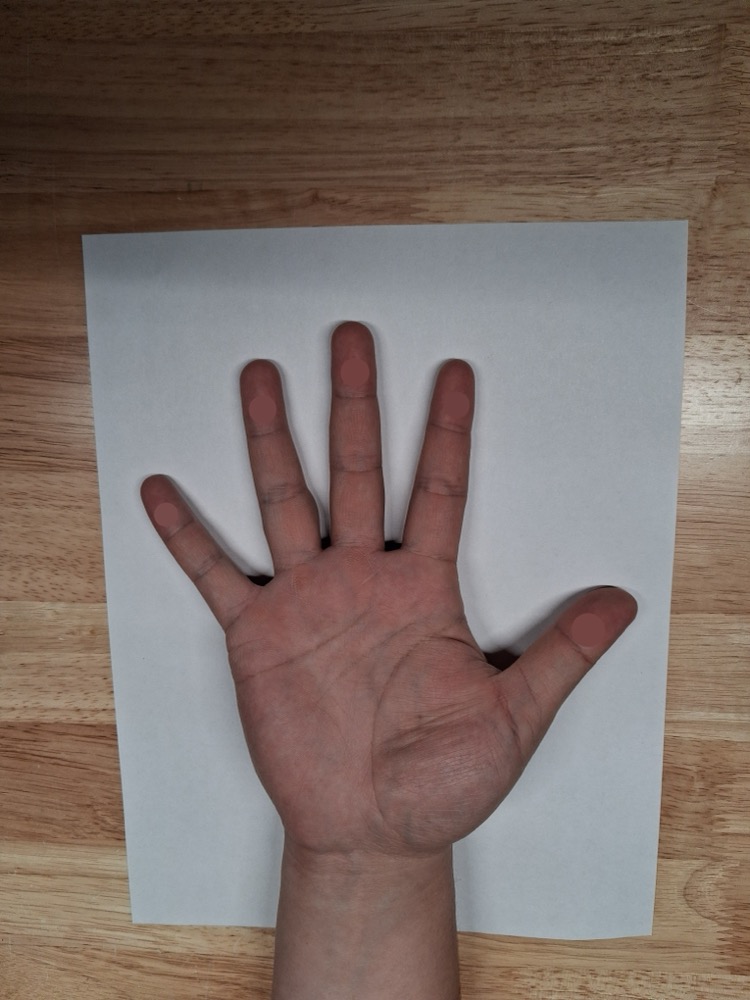}}\hfil
\subcaptionbox{Android, complex background, top-down\label{cond:d}}[0.23\textwidth]{\includegraphics[width=0.18\textwidth]{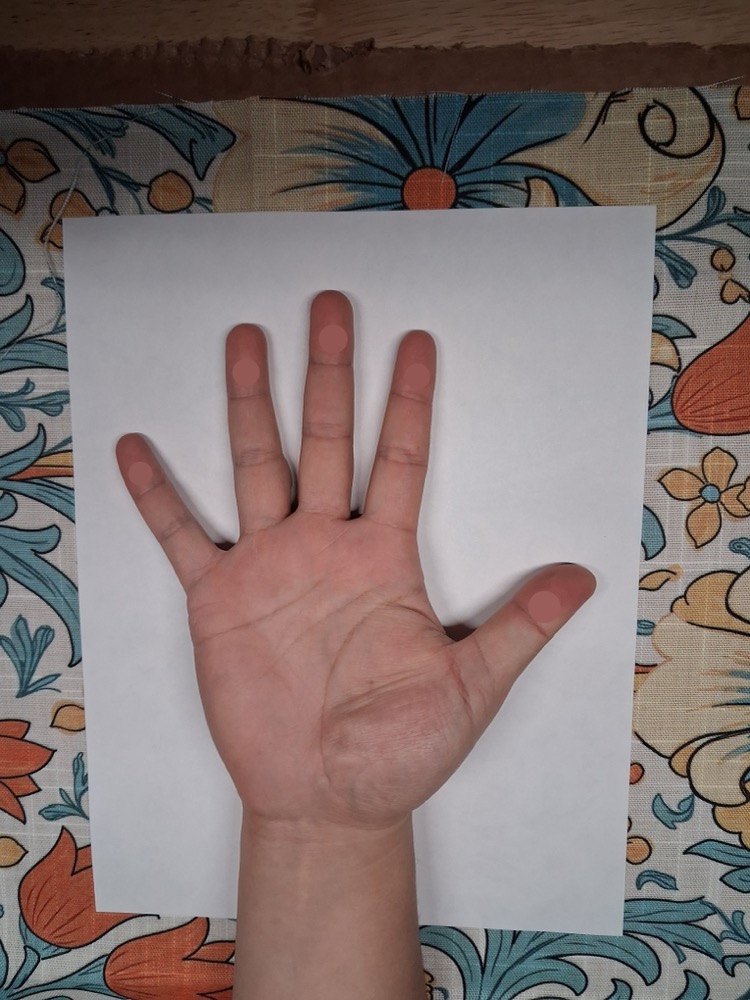}}\\[4pt]
\subcaptionbox{iPhone, simple background, oblique\label{cond:e}}[0.23\textwidth]{\includegraphics[width=0.18\textwidth]{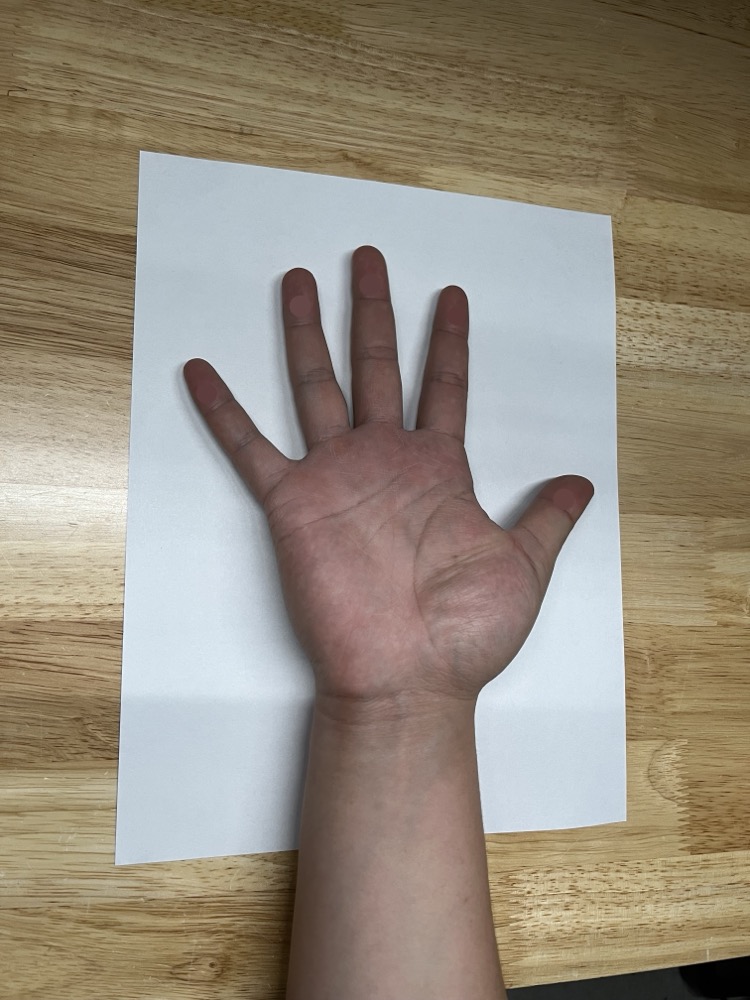}}\hfil
\subcaptionbox{iPhone, complex background, oblique\label{cond:f}}[0.23\textwidth]{\includegraphics[width=0.18\textwidth]{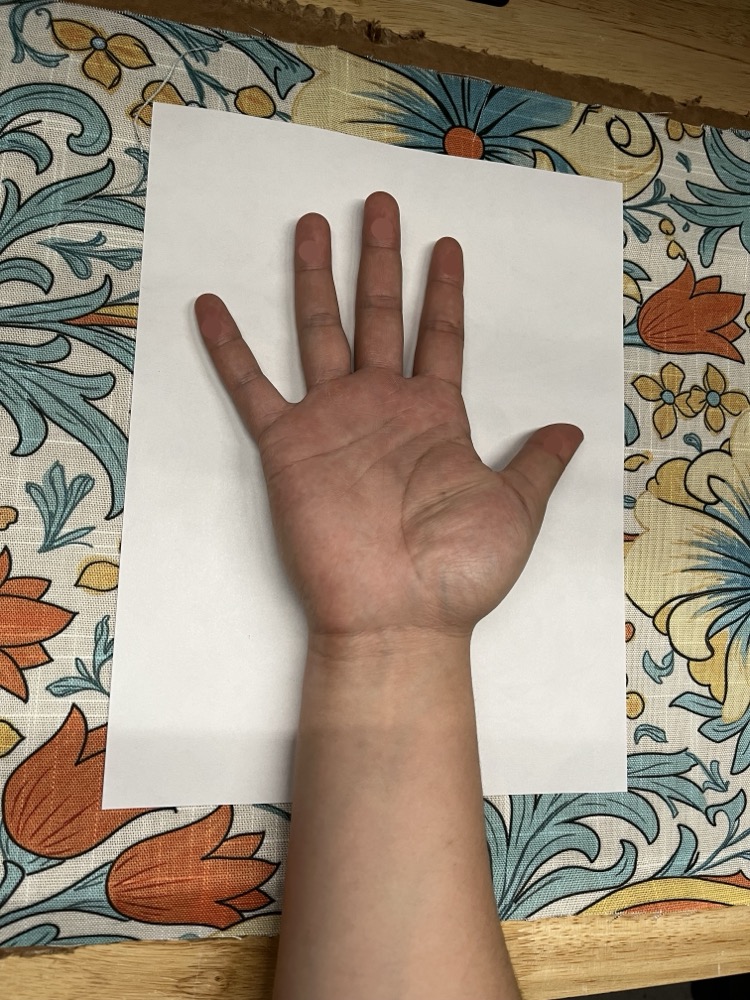}}\hfil
\subcaptionbox{Android, simple background, oblique\label{cond:g}}[0.23\textwidth]{\includegraphics[width=0.18\textwidth]{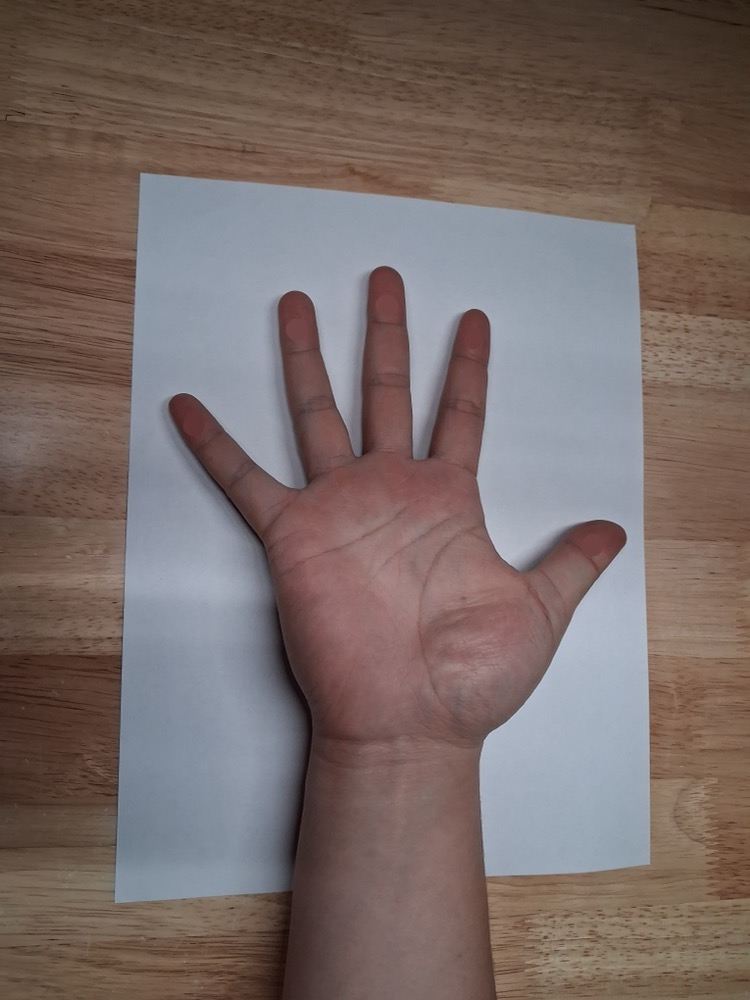}}\hfil
\subcaptionbox{Android, complex background, oblique\label{cond:h}}[0.23\textwidth]{\includegraphics[width=0.18\textwidth]{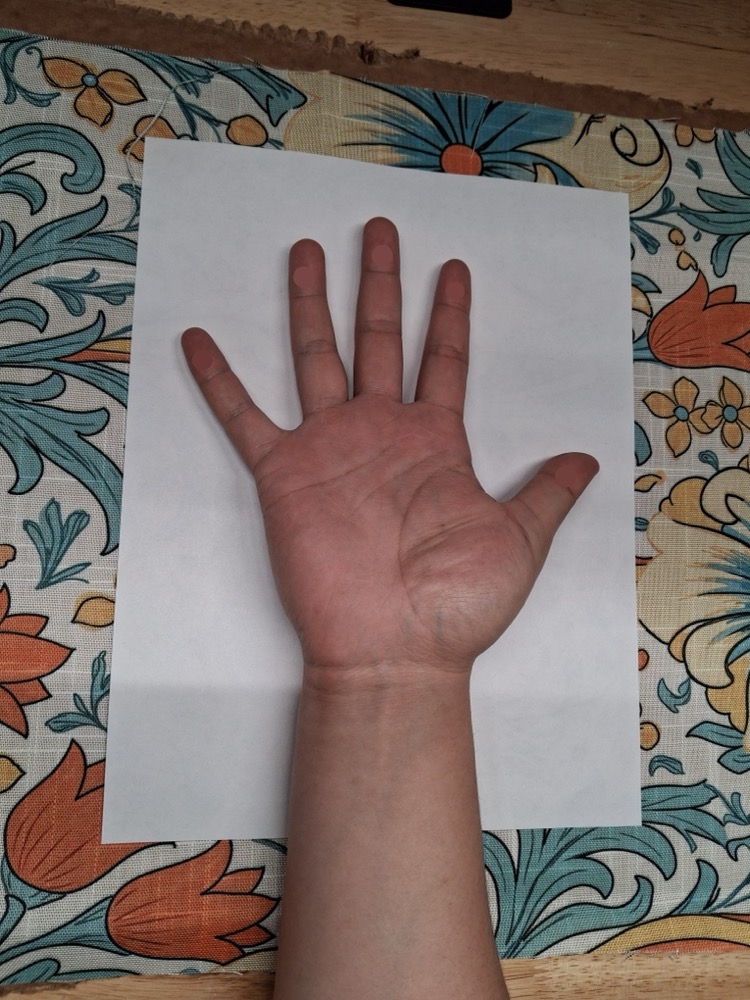}}
\caption{One author's hand under the eight test-set conditions (two devices, two backgrounds, and two capture angles; illumination pooled). Fingerprints are removed.}
\label{fig:capture_conditions}
\end{figure*}

Figure~\ref{fig:capture_conditions} illustrates the eight controlled conditions spanning two smartphones, two backgrounds, and two capture angles.

\section{Capture Rig Geometry}
\label{app:capture_rig}
\begin{figure}[!htbp]
    \centering
    \subcaptionbox{Physical rig.\label{fig:capture_rig_photo}}[0.48\columnwidth]{\includegraphics[width=0.48\columnwidth]{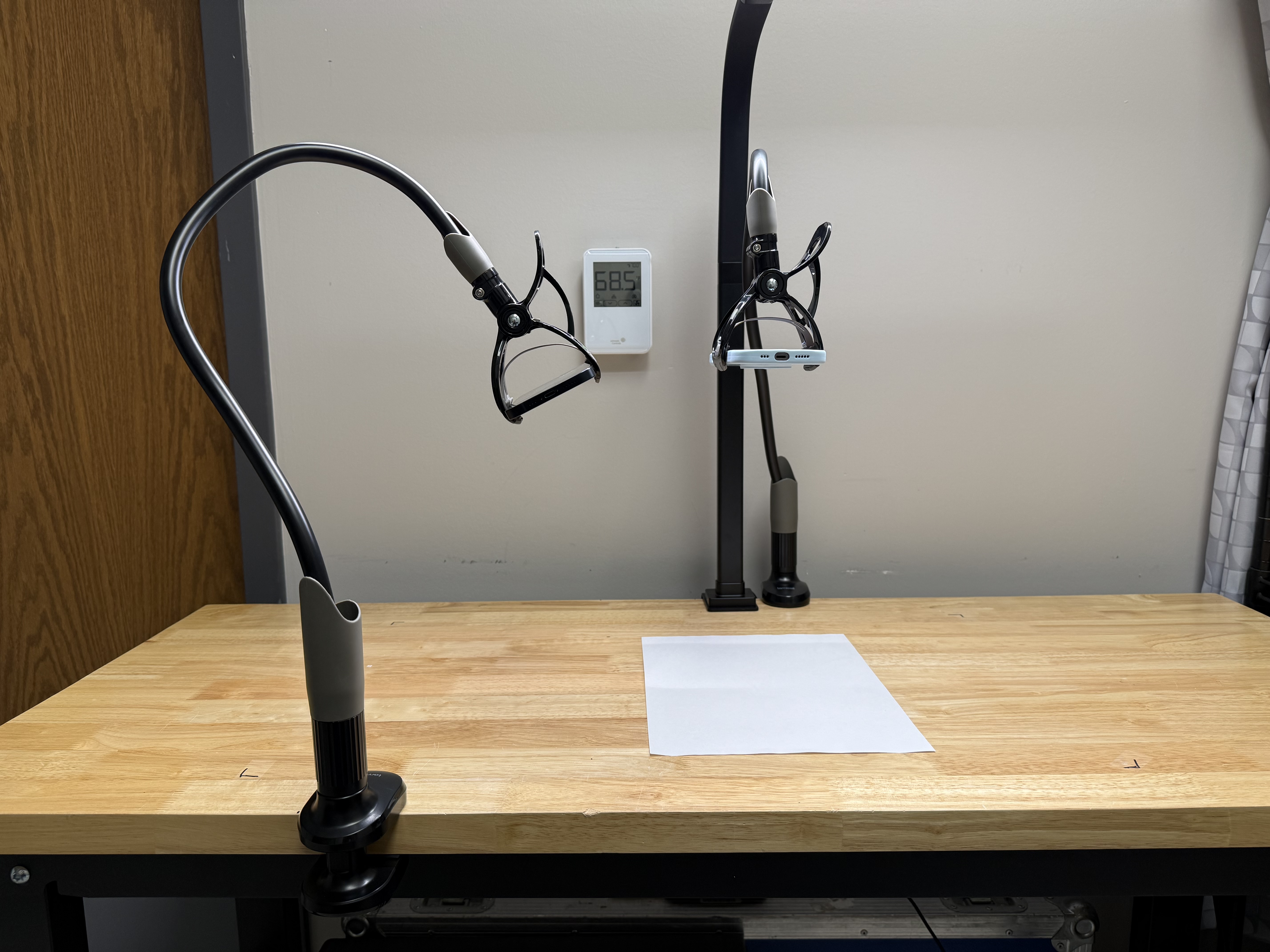}}
    \hfill
    \subcaptionbox{Front-view geometry.\label{fig:capture_rig_schematic}}[0.48\columnwidth]{\includegraphics[width=0.48\columnwidth]{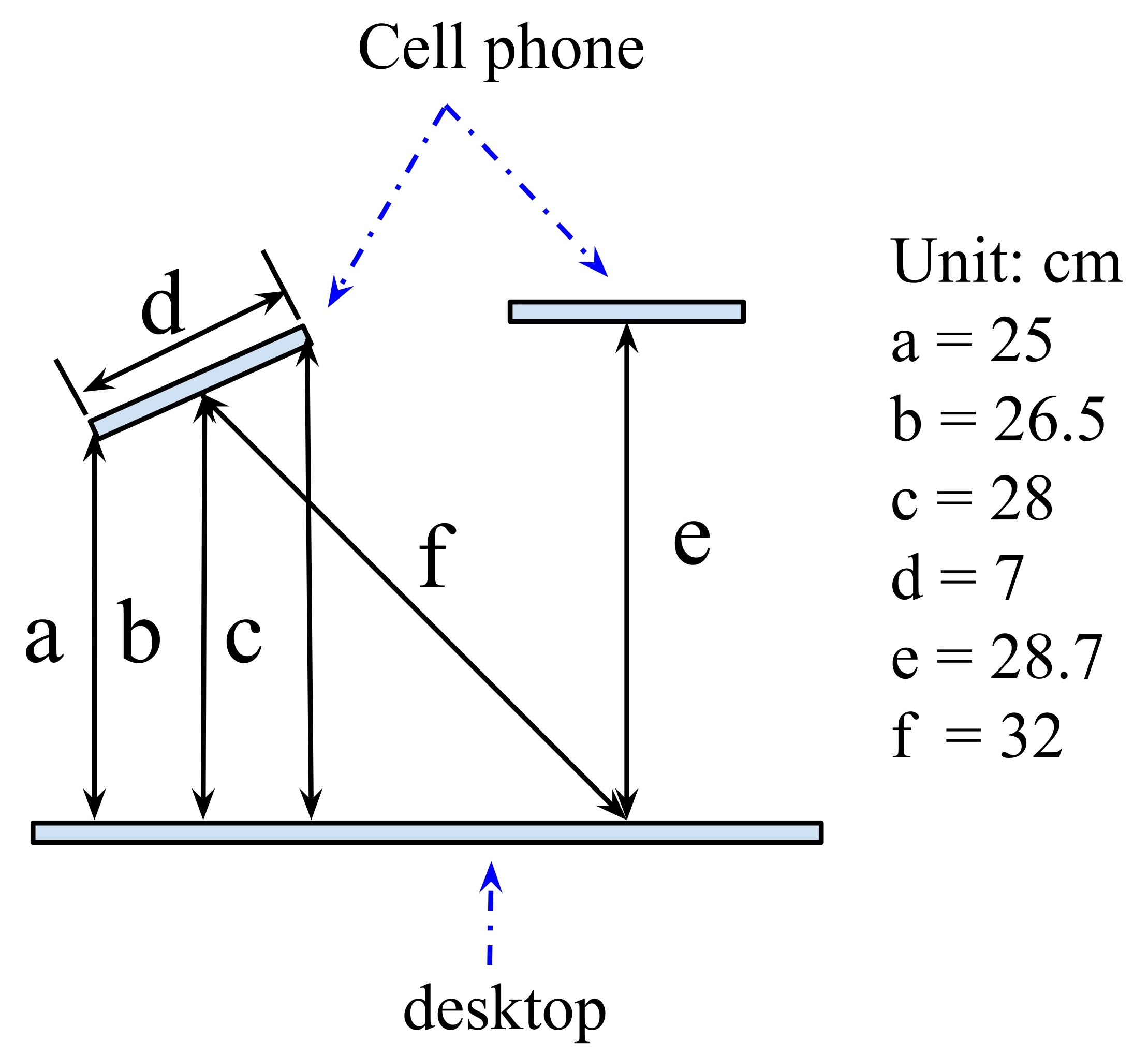}}
    \caption{Controlled capture rig: the left clamp provides the top-down view and the right clamp the oblique view. Distances are in centimeters; the measured oblique configuration is approximately $25^{\circ}$ from vertical.}
    \label{fig:capture_rig}
\end{figure}

Figure~\ref{fig:capture_rig} shows the two-holder capture rig and its measured front-view geometry.

The top-down clamp holds the phone's bottom edge 28.7\,cm above the sheet. For the oblique clamp, the near and far ends are 25 and 28\,cm above the sheet across a 7\,cm phone width, giving a tilt of $\arcsin(3/7)\approx25^{\circ}$ from vertical (Fig.~\ref{fig:capture_rig}). Phones were manually re-mounted between shots; one representative top-down mounting was tilted about $5^{\circ}$, and individual image angles were not measured.

\section{Anthropometric Landmark Definitions}
\label{app:keypoints}

\begin{figure}[!htbp]
    \centering
    \includegraphics[width=0.6\columnwidth]{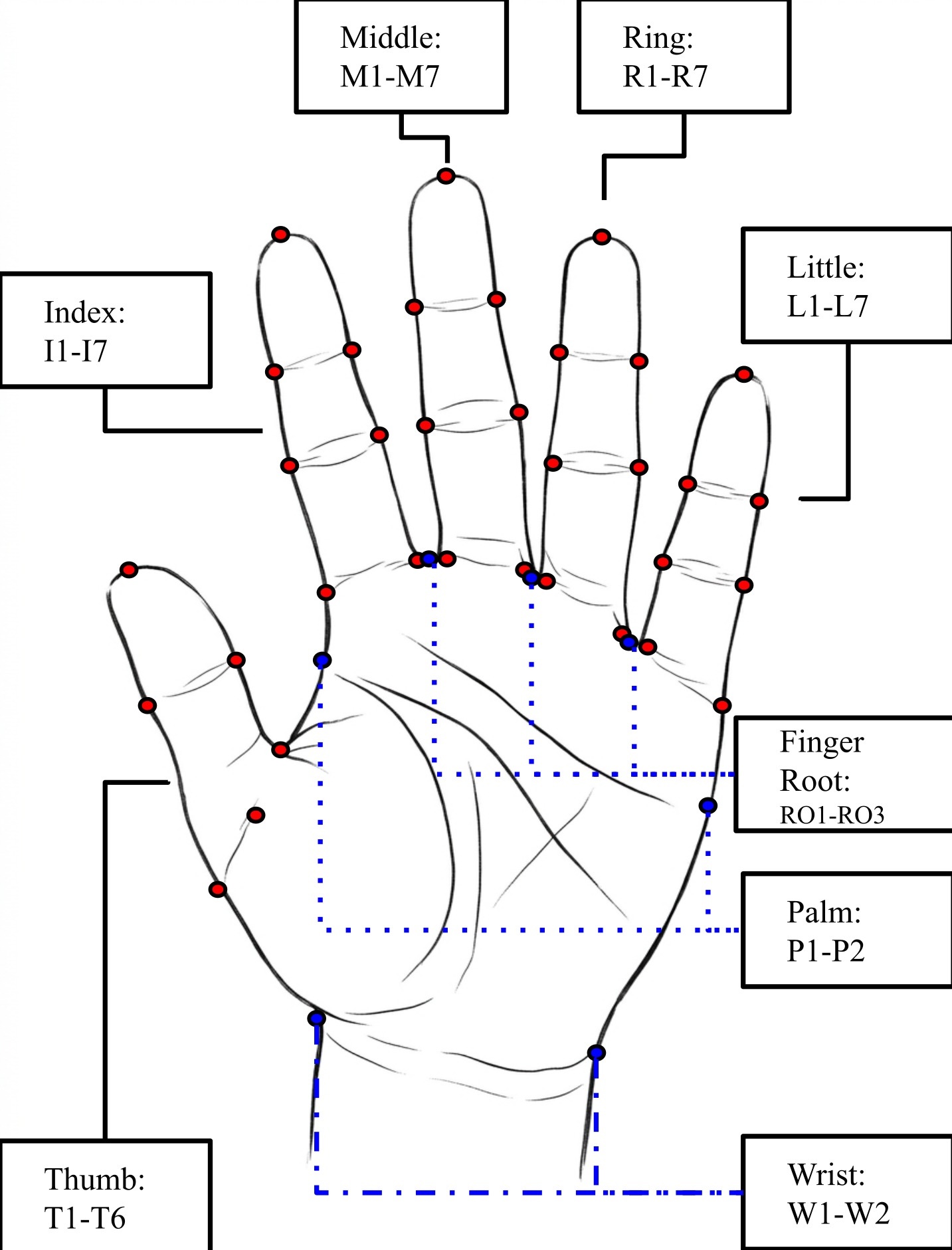}
    \caption{The 41 anthropometry-specific landmarks on a palmar hand, labeled by abbreviation (Table~\ref{tab:keypoint_groups}).}
    \label{fig:keypoints}
\end{figure}

HandAnthro predicts 41 anthropometry-specific 2D landmarks on each palmar hand image. Fig.~\ref{fig:keypoints} shows them on a representative hand, and Table~\ref{tab:keypoint_groups} gives their per-region names. From these 41 directly predicted landmarks, the metric-conversion stage deterministically derives 14 midpoint landmarks (Table~\ref{tab:derived_landmarks}). For landmarks $k_a=(x_a,y_a)$ and $k_b=(x_b,y_b)$, each derived midpoint is
\begin{equation}
m(a,b)=\frac{k_a+k_b}{2}
=\left(\frac{x_a+x_b}{2},\frac{y_a+y_b}{2}\right).
\label{eq:midpoint}
\end{equation}

The submitted implementation contains exactly 44 endpoint pairs, listed with their anatomical names and errors in Table~\ref{tab:dim_full}.

\begin{table*}[!tp]
    \centering
    \footnotesize
    \setlength{\tabcolsep}{6pt}
    \begin{tabular}{llcc}
    \toprule
    Region & Abbreviation range & Count & Role \\
    \midrule
    Thumb                       & $T_1$ to $T_6$                                  & 6  & Thumb-specific landmarks \\
    Index finger                & $I_1$ to $I_7$                                  & 7  & Fingertip to finger root \\
    Middle finger               & $M_1$ to $M_7$                                  & 7  & Fingertip to finger root \\
    Ring finger                 & $R_1$ to $R_7$                                  & 7  & Fingertip to finger root \\
    Little finger               & $L_1$ to $L_7$                                  & 7  & Fingertip to finger root \\
    Inter-finger webbing        & $\mathrm{RO}_1$ to $\mathrm{RO}_3$              & 3  & Root-of-finger webbing points \\
    Wrist / palm                & $W_1$, $W_2$, $P_1$, $P_2$                    & 4  & Wrist and palm reference points \\
    \midrule
    \textbf{Total}              &                                               & \textbf{41} &  \\
    \bottomrule
    \end{tabular}
    \caption{The 41 anthropometry-specific landmarks predicted directly by the YOLO model, grouped by anatomical region.}
    \label{tab:keypoint_groups}
\end{table*}

\begin{table*}[!tp]
    \centering
    \footnotesize
    \setlength{\tabcolsep}{5pt}
    \begin{tabular}{llll}
    \toprule
    Abbrev.\ & Expansion & Anatomical joint (palmar crease) & Formula \\
    \midrule
    $\mathrm{MFK}_I$ & Mid First Knuckle (Index)   & Index-finger DIP crease  & midpoint($I_2$, $I_3$) \\
    $\mathrm{MFK}_M$ & Mid First Knuckle (Middle)  & Middle-finger DIP crease & midpoint($M_2$, $M_3$) \\
    $\mathrm{MFK}_R$ & Mid First Knuckle (Ring)    & Ring-finger DIP crease   & midpoint($R_2$, $R_3$) \\
    $\mathrm{MFK}_L$ & Mid First Knuckle (Little)  & Little-finger DIP crease & midpoint($L_2$, $L_3$) \\
    $\mathrm{MFK}_T$ & Mid First Knuckle (Thumb)   & Thumb IP crease          & midpoint($T_2$, $T_3$) \\
    \midrule
    $\mathrm{MSK}_I$ & Mid Second Knuckle (Index)  & Index-finger PIP crease  & midpoint($I_4$, $I_5$) \\
    $\mathrm{MSK}_M$ & Mid Second Knuckle (Middle) & Middle-finger PIP crease & midpoint($M_4$, $M_5$) \\
    $\mathrm{MSK}_R$ & Mid Second Knuckle (Ring)   & Ring-finger PIP crease   & midpoint($R_4$, $R_5$) \\
    $\mathrm{MSK}_L$ & Mid Second Knuckle (Little) & Little-finger PIP crease & midpoint($L_4$, $L_5$) \\
    \midrule
    $\mathrm{MTK}_I$ & Mid Third Knuckle (Index)   & Index-finger MCP crease  & midpoint($I_6$, $I_7$) \\
    $\mathrm{MTK}_M$ & Mid Third Knuckle (Middle)  & Middle-finger MCP crease & midpoint($M_6$, $M_7$) \\
    $\mathrm{MTK}_R$ & Mid Third Knuckle (Ring)    & Ring-finger MCP crease   & midpoint($R_6$, $R_7$) \\
    $\mathrm{MTK}_L$ & Mid Third Knuckle (Little)  & Little-finger MCP crease & midpoint($L_6$, $L_7$) \\
    \midrule
    $\mathrm{MoW}$   & Midpoint of Wrist                & Midpoint of wrist line   & midpoint($W_1$, $W_2$) \\
    \bottomrule
    \end{tabular}
    \caption{The 14 midpoint landmarks derived from the 41 directly predicted anthropometry-specific landmarks.}
    \label{tab:derived_landmarks}
\end{table*}

\section{Adaptive SAM-HQ Prompts}
\label{app:samhq_prompts}
The PR stage uses separate SAM-HQ calls to produce the paper mask $M_{\text{paper}}$ and raw hand mask $M_{\text{hand-raw}}$. Positive prompts mark the target region; negative prompts mark regions to exclude from that mask. Let $m_j=(x_j,y_j)$, $j=0,\ldots,20$, denote the MediaPipe landmarks in image coordinates, with $x$ increasing rightward and $y$ downward. Here $m_0$, $m_9$, and $m_{12}$ locate the wrist, middle-finger root, and middle fingertip, respectively. The middle-finger root-to-tip Euclidean distance, $L=\lVert m_{12}-m_9\rVert_2$, sets the scale of the prompt offsets.

\begin{figure}[!htbp]
    \centering
    \subcaptionbox{Paper-mask prompts.\label{app:fig:samhq_prompts_paper}}[0.48\columnwidth]{\includegraphics[width=0.48\columnwidth]{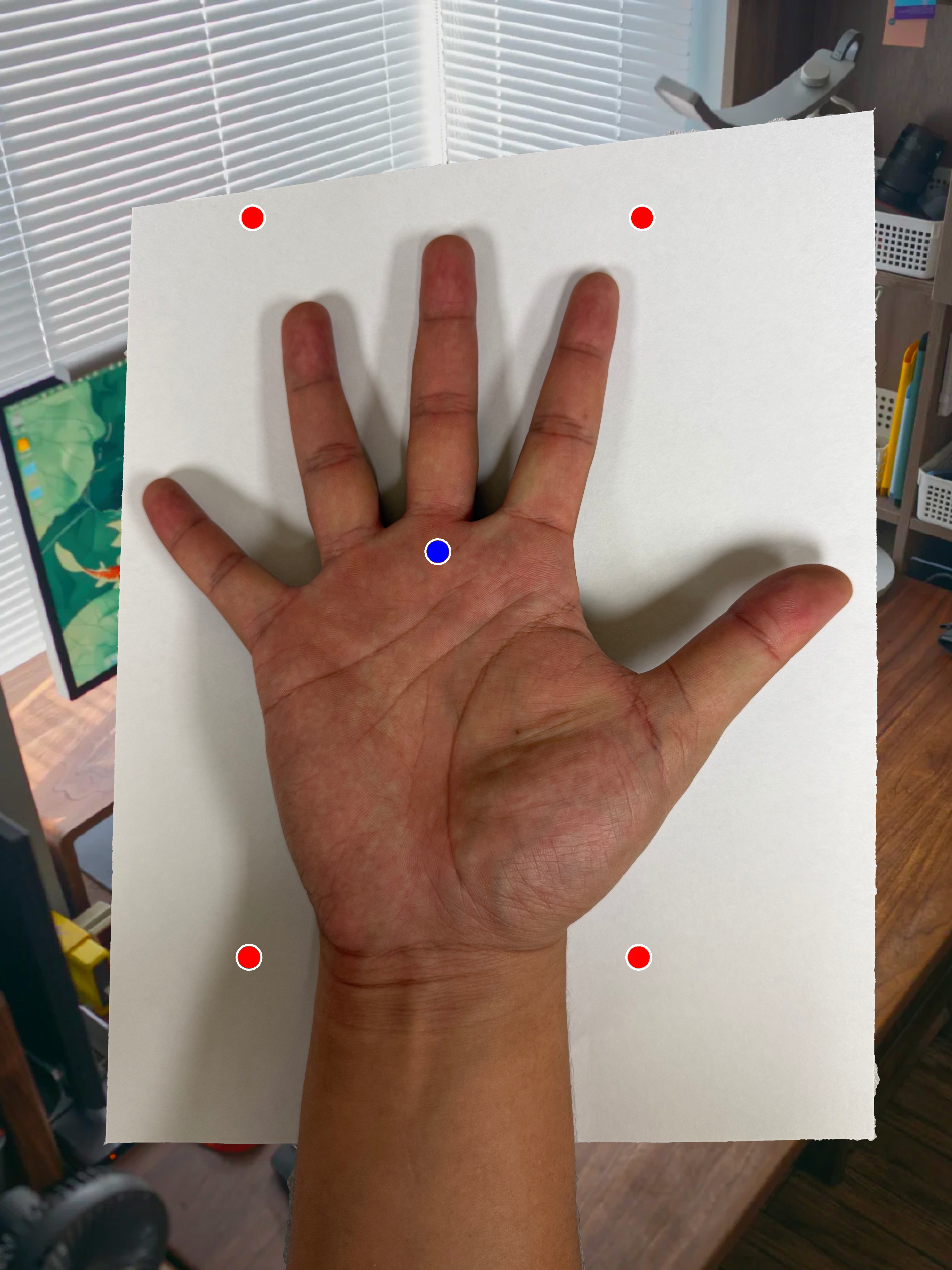}}\hfill
    \subcaptionbox{Hand-mask prompts.\label{app:fig:samhq_prompts_hand}}[0.48\columnwidth]{\includegraphics[width=0.48\columnwidth]{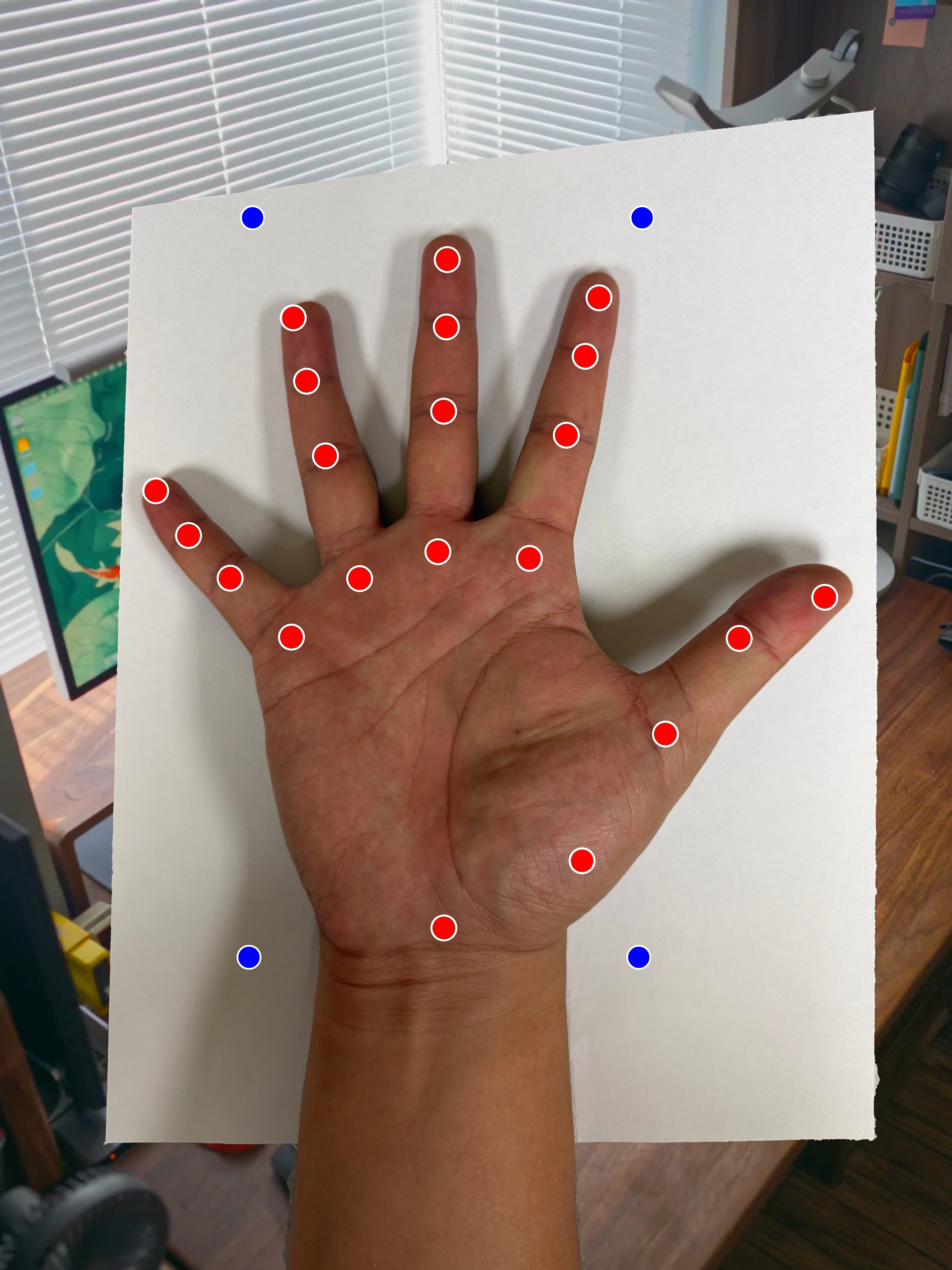}}
    \caption{Illustrative SAM-HQ prompts for (a) paper-mask and (b) hand-mask generation. Red points are positive prompts; blue points are negative prompts.}
    \label{app:fig:samhq_prompts}
\end{figure}

\paragraph{Paper-mask prompts.} The four positive prompts are labeled top left (TL), top right (TR), bottom left (BL), and bottom right (BR):
\begin{equation}
\begin{aligned}
p_{\mathrm{TL/TR}}&=\operatorname{clip}\!\left(m_{12}+\left(\mp\frac{L}{1.5},-\frac{L}{7}\right)\right),\\
p_{\mathrm{BL/BR}}&=\operatorname{clip}\!\left(m_{0}+\left(\mp\frac{L}{1.5},\phantom{-}\frac{L}{10}\right)\right),
\end{aligned}
\label{eq:paper_prompts}
\end{equation}
The upper pair starts at the middle fingertip $m_{12}$, moves $L/1.5$ left or right, and $L/7$ upward. The lower pair starts at the wrist $m_0$, uses the same horizontal offset, and moves $L/10$ downward. In $\mp$, the minus sign gives TL/BL and the plus sign gives TR/BR; $\operatorname{clip}$ keeps each point within the image bounds. The single negative prompt $n=m_9$ marks the hand interior for exclusion from the paper mask. No box prompt is used. Figure~\ref{app:fig:samhq_prompts_paper} shows the paper prompts.

\paragraph{Hand-mask prompts.} Let $\mathcal P$ and $\mathcal H$ denote the paper and hand prompt sets, with superscripts $+$ and $-$ indicating positive and negative labels. Define $\mathcal P^{+}=\{p_{\mathrm{TL}},p_{\mathrm{TR}},p_{\mathrm{BL}},p_{\mathrm{BR}}\}$ from Eq.~\ref{eq:paper_prompts}. Hand segmentation uses all 21 MediaPipe landmarks as positive prompts and reuses these four paper locations as negative prompts:
\begin{equation}
\begin{aligned}
\mathcal P^{-}&=\{m_9\},\\
\mathcal H^{+}&=\{m_0,\ldots,m_{20}\},\qquad
\mathcal H^{-}=\mathcal P^{+}.
\end{aligned}
\label{eq:shared_sam_prompts}
\end{equation}
Before LaMa inpainting, $M_{\text{hand-raw}}$ is dilated once using an all-ones $31\times31$ square structuring element (15-pixel half-width). This expanded mask is used only for the LaMa inpainting region; BW later reuses the undilated hand mask.

\section{Quadrilateral Detection and Homography}
\label{app:quad_detection_params}

The PR stage recovers the paper quadrilateral from the inpainting-completed paper mask $M_{\text{paper-com}}$ using the following fixed parameters. The mask is smoothed with an $11\times11$ Gaussian kernel, then processed with Canny edge detection~\cite{canny1986computational} using thresholds $50$ and $150$. Morphological closing uses a $3\times3$ rectangular kernel for three dilation and erosion iterations. The five largest contours are approximated using Douglas--Peucker simplification~\cite{douglas1973algorithms} with tolerance $\varepsilon=0.08\,L_{\text{peri}}$, where $L_{\text{peri}}$ is the contour perimeter; the first candidate with exactly four vertices and area greater than 30\% of the image is accepted.

Let the accepted vertices, ordered top-left, top-right, bottom-right, bottom-left, be $q_i$, and let $q_i'$ be the corresponding corners of the destination rectangle. Its width and height are initialized from the larger opposing side lengths; then the smaller adjustment required to enforce the letter-paper ratio $H'/W'=11/8.5$ is applied. The homography $\mathbf H$ is the projective map satisfying
\begin{equation}
\begin{aligned}
\lambda_i
\begin{bmatrix}q'_{i,x}&q'_{i,y}&1\end{bmatrix}^{\!T}
&=\mathbf H
\begin{bmatrix}q_{i,x}&q_{i,y}&1\end{bmatrix}^{\!T},\\[-2pt]
&\hspace{-4em} i=1,\ldots,4.
\end{aligned}
\label{eq:homography}
\end{equation}
The same $\mathbf H$ warps the original RGB image and the undilated hand mask, preserving their pixel correspondence. Immediately before landmark inference, both rectified outputs are resized to the canonical $720\times932$\,px model frame used for metric conversion (Section~\ref{sec:scale_estimation}).

\section{Background Whitening}
\label{app:bw_masks}

\begin{figure}[!htbp]
\centering
\includegraphics[width=\columnwidth]{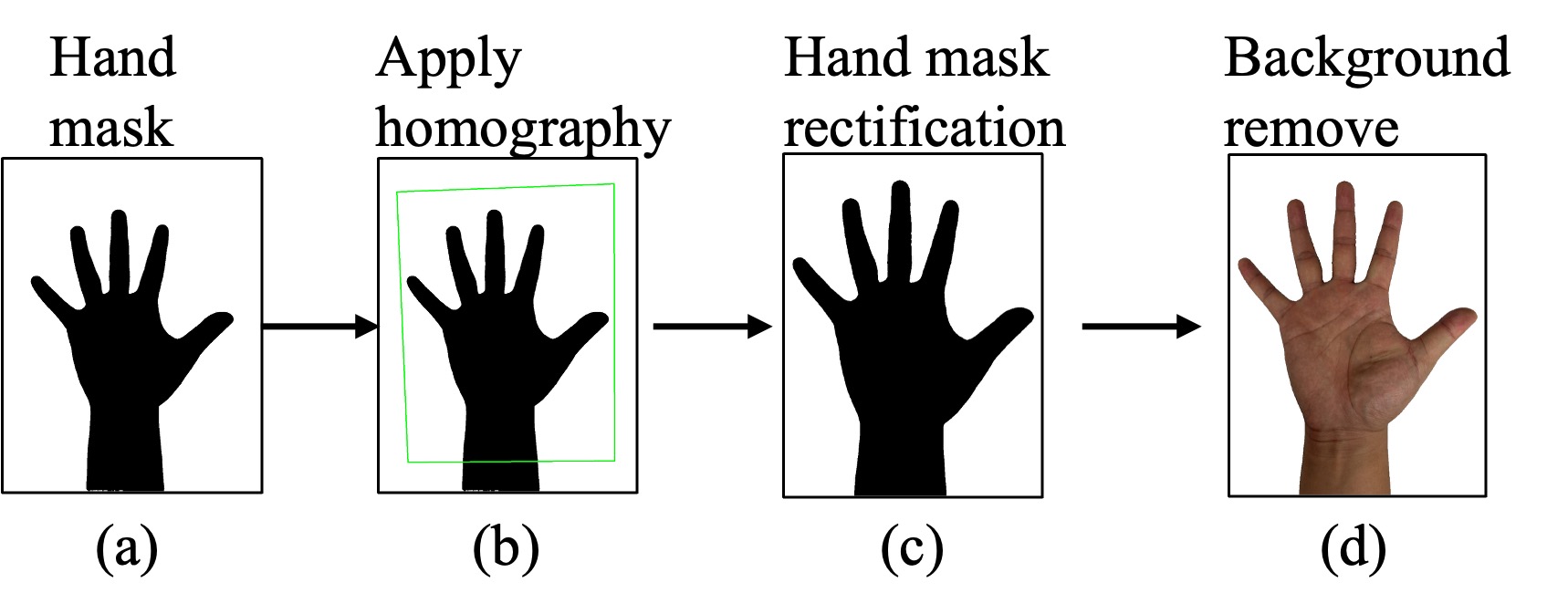}
\caption{Background whitening using the warped hand mask: (a) raw hand mask; (b) detected paper quadrilateral (green); (c) rectified hand mask; (d) background-whitened image.}
\label{Fig.BW}
\end{figure}

Fig.~\ref{Fig.BW} illustrates background whitening (BW). The RGB image and undilated hand mask are warped using the same homography to keep them aligned. The stored mask represents the hand in black; after warping, a pixel is treated as hand foreground if at least one mask channel has a value below 100. BW preserves the rectified image's RGB values at these foreground pixels and replaces all remaining pixels with white (255 in each RGB channel).

\section{YOLO Fine-Tuning Hyperparameters}
\label{app:hparams}

Table~\ref{tab:yolo_key_hparams_appendix} lists the key YOLO-Pose fine-tuning settings used for all 16 configurations.

\begin{table}[!htbp]
  \centering
  \footnotesize
  \setlength{\tabcolsep}{4pt}
  \begin{tabularx}{\columnwidth}{@{}l>{\raggedright\arraybackslash}X@{}}
    \toprule
    \textbf{Item} & \textbf{Value} \\
    \midrule
    Epochs / batch size & 8000 / 40 \\
    Early stopping patience & 10000 \\
    Compute & CUDA GPU; dataloader workers: 8 \\
    Optimizer & \texttt{auto} (Ultralytics default selection) \\
    Learning rate schedule & $\mathrm{lr}_0=0.001$, $\mathrm{lrf}=0.01$ \\
    Momentum / weight decay & 0.937 / $5\times10^{-4}$ \\
    Warmup & 3 epochs (\texttt{warmup\_epochs}=3.0) \\
    Regularization & dropout = 0.9 \\
    Loss weights & box = 0.2,\; cls = 0.1,\; pose = 20.0,\; kobj = 3.0 \\
    Augmentation (selected) &
      translate = 0.1,\; scale = 0.5,\; fliplr = 0.5,\; mosaic = 1.0,\;
      \texttt{auto\_}\allowbreak\texttt{augment} = randaugment,\; erasing = 0.4 \\
    Reproducibility & seed = 0,\; deterministic = True,\; AMP = True \\
    \bottomrule
  \end{tabularx}
  \caption{Key hyperparameters and settings for YOLO fine-tuning.}
  \label{tab:yolo_key_hparams_appendix}
\end{table}

\section{Geometry-Constrained Post-Processing}
\label{app:pp_formulation}

PP runs MediaPipe Hands on $I_{\text{white}}$ to obtain 21 auxiliary landmarks $\mathcal K^{MP}$, then refines the 41 YOLO landmarks $\mathcal K^{Y}$ using contour searches on the same image. It constructs separation lines, estimates finger axes and fingertips, applies per-finger similarity transforms, and refines boundary endpoints in that order. A valid boundary pixel is non-white, has at least three non-white pixels in its $3\times3$ neighborhood, and has at least one white neighbor. The per-channel white threshold is 250 for finger-axis estimation and 230 for subsequent refinement.

\begin{figure}[!htbp]
\centering
\includegraphics[width=0.70\columnwidth]{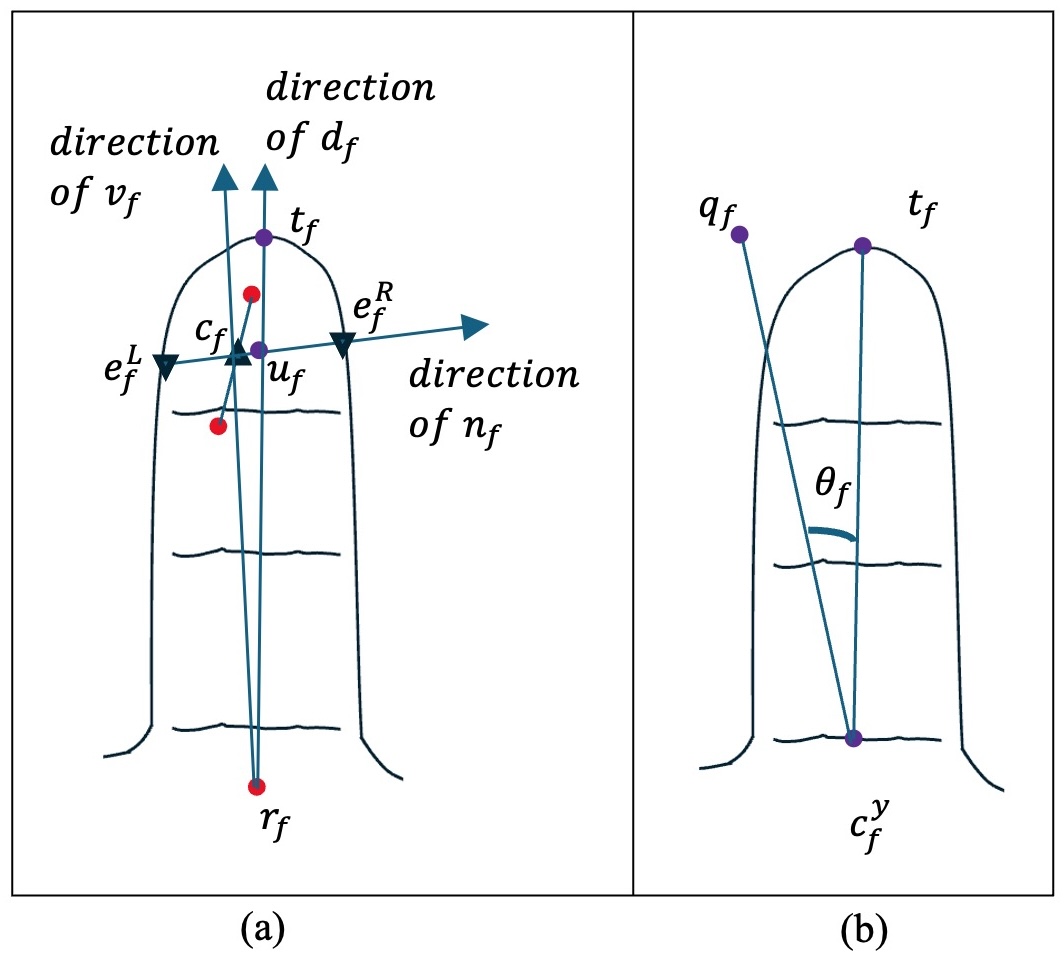}
\caption{Finger-axis estimation and fingertip correction for finger $f$.}
\label{Visualization of finger axis and correct finger tip point estimation}
\end{figure}

\paragraph{Separation lines and finger axes.} Three separation lines, index--middle, middle--ring, and ring--little, connect each neighboring pair's MediaPipe MCP midpoint to its PIP midpoint. They stop boundary-search rays from crossing into adjacent fingers; the thumb has no separation constraint. For finger $f$, let $r_f$ be its MediaPipe root and $c_f$ the midpoint of its fingertip and adjacent tip-side joint. Search from $c_f$ perpendicular to $c_f-r_f$ in both directions for the first boundary pixels $e_f^L,e_f^R$. Their midpoint and the corrected axis are
\begin{equation}
u_f=\frac{e_f^L+e_f^R}{2},\qquad
d_f=\frac{u_f-r_f}{\|u_f-r_f\|_2}.
\end{equation}
A forward search from $u_f$ along $d_f$ finds the corrected fingertip $t_f$ (Fig.~\ref{Visualization of finger axis and correct finger tip point estimation}).

\paragraph{Per-finger similarity transform.} Let $q_f$ and $c_f^y$ be the YOLO fingertip and fixed root. The signed angle $\theta_f$ rotates $q_f-c_f^y$ onto $t_f-c_f^y$. With $R(\theta_f)$ the 2D rotation matrix, each member $k_i$ of finger group $G_f$ is updated by
\begin{equation}
\begin{aligned}
s_f&=\frac{\|t_f-c_f^y\|_2}{\|q_f-c_f^y\|_2},\\
k_i'&=c_f^y+s_fR(\theta_f)(k_i-c_f^y).
\end{aligned}
\end{equation}

\paragraph{Boundary snapping and special-point refinement.} From each endpoint, search bidirectionally along its current configured measurement-pair line for the nearest valid boundary within $0.03W$ (21.6\,px at $W=720$), checking a $3\times3$ neighborhood at each sampled position. Exclude fingertips, derived midpoints (including $\mathrm{MoW}$), $T_5$, $T_6$, and $\mathrm{RO}_{1:3}$ from this bounded snap. Unresolved wrist endpoints use an uncapped search along the wrist line, inward on white pixels and outward on hand pixels. Recompute $\mathrm{MoW}$, then check $\mathrm{RO}_{1:3}$ against neighboring finger-root boundaries and move them toward $\mathrm{MoW}$ when needed. If $T_5$ lies on a white pixel, search from it only toward $\mathrm{MoW}$; subsequently correct an off-boundary $T_4$ by uncapped directional search along $T_5\!\rightarrow\!T_4$. Correct any off-boundary $T_2$/$T_3$ by uncapped search along their incoming pair axis, outward from the other endpoint on hand pixels and inward toward it on white pixels. Recompute all midpoint landmarks (Eq.~\ref{eq:midpoint}) and the 44 metric dimensions (Section~\ref{sec:scale_estimation}) after refinement.

\section{Stage-Level Comparison Methods}
\label{app:baseline_details}
This section documents the alternatives substituted for HandAnthro's PR and BW stages.

\subsection{Canny--Hough Perspective Rectification (CH-PR)}
\label{app:baseline_prcv}
We implement a fully automatic, learning-free baseline for paper rectification using edge and line geometry. Given an input image, we compute a Canny edge map~\cite{canny1986computational} and extract line segments via the probabilistic Hough transform~\cite{matas2000robust} (OpenCV \texttt{HoughLinesP}). Segments are clustered into two dominant orientation groups and assigned to the four paper sides, and each side line is fitted from its supporting segments. We select the candidate with maximal edge support on the Canny map. Paper corners are computed from intersections of adjacent side lines, and the quadrilateral is required to be convex. The image is then warped to a canonical rectangle via homography.

\paragraph{Failure-mode breakdown.}
On the 720-image test set, CH-PR returned 412 convex, warpable quadrilaterals (57.2\%), with downstream processing completed for all corresponding captures. MAE on the 406 captures completed by both pipelines was 10.68\,mm for CH-PR and 3.66\,mm for HandAnthro. The 308 non-warpable cases comprised 266 without two orthogonal line groups, 32 with non-convex side-line intersections, and 10 with an unrecoverable missing side.

\subsection{Background Removal Baselines (REMBG, BackgroundRemover, and CarveKit)}
\label{app:baseline_bg}
\begin{figure*}[!tp]
    \centering
    \includegraphics[width=0.92\textwidth]{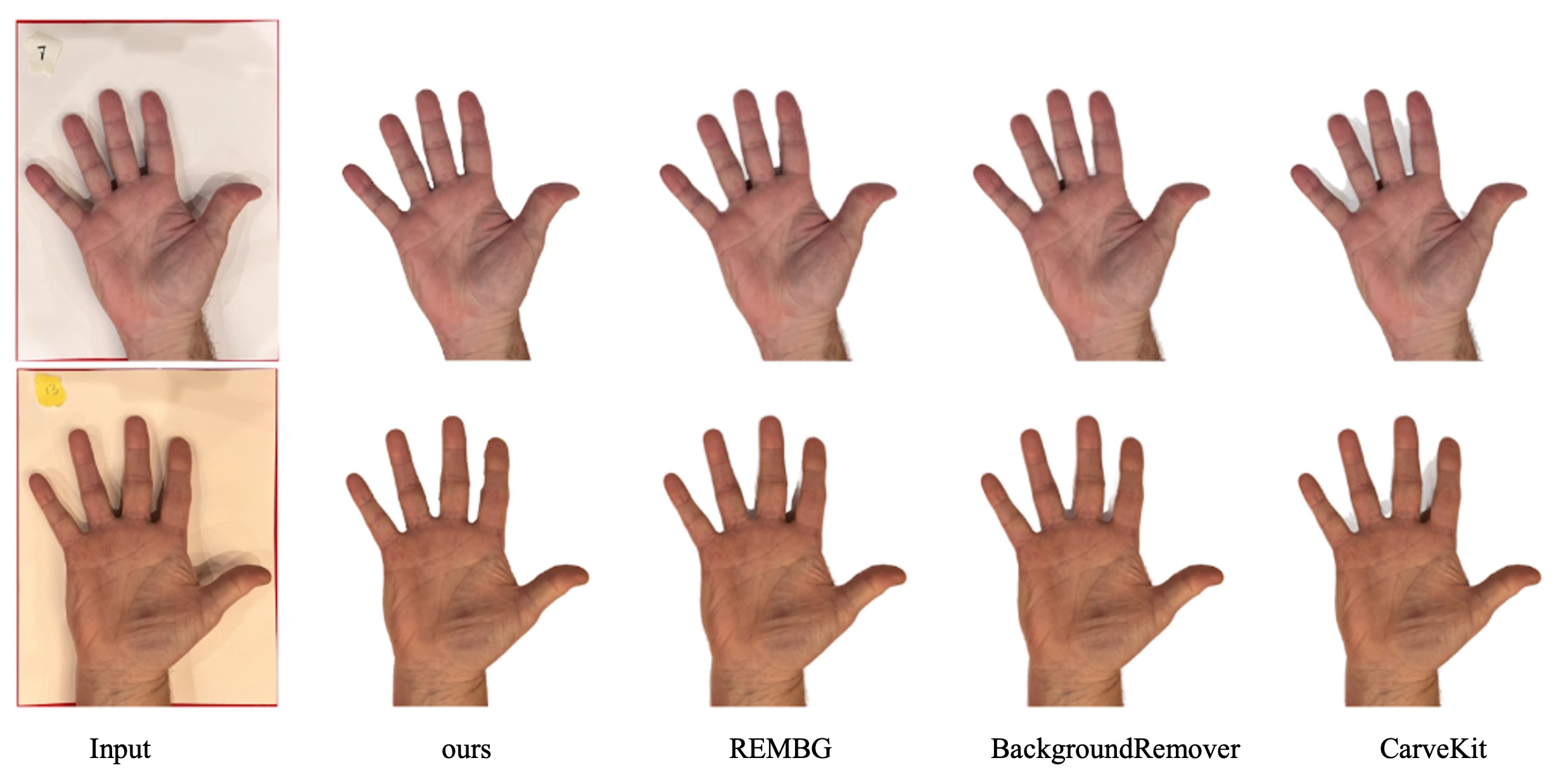}
    \caption{Our BW versus the three off-the-shelf baselines on representative images. REMBG, BackgroundRemover, and CarveKit often retain inter-finger shadows that blur the hand boundary, whereas our BW gives a clean separation.}
    \label{Fig:BW qualitative}
\end{figure*}

To evaluate the accuracy and computational efficiency of our BW module, we compare it with REMBG~\cite{Gatis_rembg_2025}, BackgroundRemover~\cite{nader_backgroundremover_pypi}, and CarveKit~\cite{selin2024carvekit_github} under identical PR, landmark-prediction, and PP stages. Table~\ref{tab:bw_comparison} reports end-to-end dimension MAE and BW-only runtime, while Fig.~\ref{Fig:BW qualitative} shows representative mask behavior under inter-finger shadows. Every variant produced complete outputs for all 704 PR-complete captures.

As an architectural fairness check, forcing our BW to run from scratch on each rectified image without reusing the PR-stage hand mask took 1213\,ms per image on the same NVIDIA RTX 4000 Ada Generation GPU, compared with 93\,ms for REMBG. The production advantage therefore comes from mask reuse rather than a faster standalone segmentation model.

\begin{table}[!htbp]
    \centering
    \footnotesize
    \setlength{\tabcolsep}{3pt}
    \begin{tabular}{@{}lcccc@{}}
    \toprule
    Variant & \makecell{Mean\\(mm)} & \makecell{Median\\(mm)} & \makecell{$\Delta$MAE\\(mm / \%)} & \makecell{Runtime\\(ms)} \\
    \midrule
    \textbf{Ours}       & \textbf{3.80} & \textbf{3.45} & ---            & \textbf{17} \\
    REMBG               & 4.26          & 3.79          & $+0.46$ / $+12.0\%$ & 93 \\
    \makecell[l]{Background\\Remover} & 4.26 & 3.82 & $+0.46$ / $+12.1\%$ & 188 \\
    CarveKit            & 4.71          & 4.48          & $+0.91$ / $+23.9\%$ & 379 \\
    \bottomrule
    \end{tabular}
    \caption{Background-whitening alternatives under identical PR, landmark-prediction, and PP stages. MAE is measured against calipers on the 704 PR-complete captures, all of which yielded complete outputs for every variant; runtime is BW-only mean latency over the same 704 images. Positive $\Delta$MAE indicates degradation relative to mask reuse.}
    \label{tab:bw_comparison}
\end{table}

\begin{figure}[!htbp]
    \centering
    \includegraphics[width=\columnwidth]{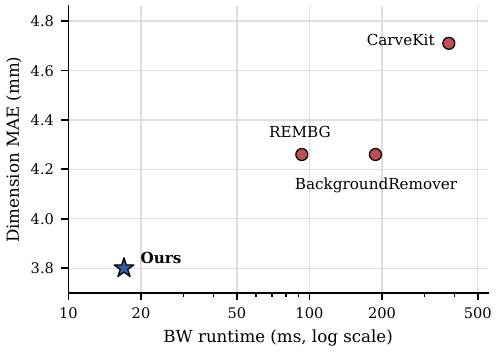}
    \caption{BW-stage runtime versus caliper-referenced dimension MAE on the 704 complete captures. Runtime is measured on an NVIDIA RTX 4000 Ada Generation GPU; the horizontal axis is logarithmic. Values match Table~\ref{tab:bw_comparison}.}
    \label{fig:bw_runtime_accuracy}
\end{figure}

Figure~\ref{fig:bw_runtime_accuracy} plots the runtime and MAE values in Table~\ref{tab:bw_comparison}.

\begin{figure}[!htbp]
\centering
\includegraphics[width=\columnwidth]{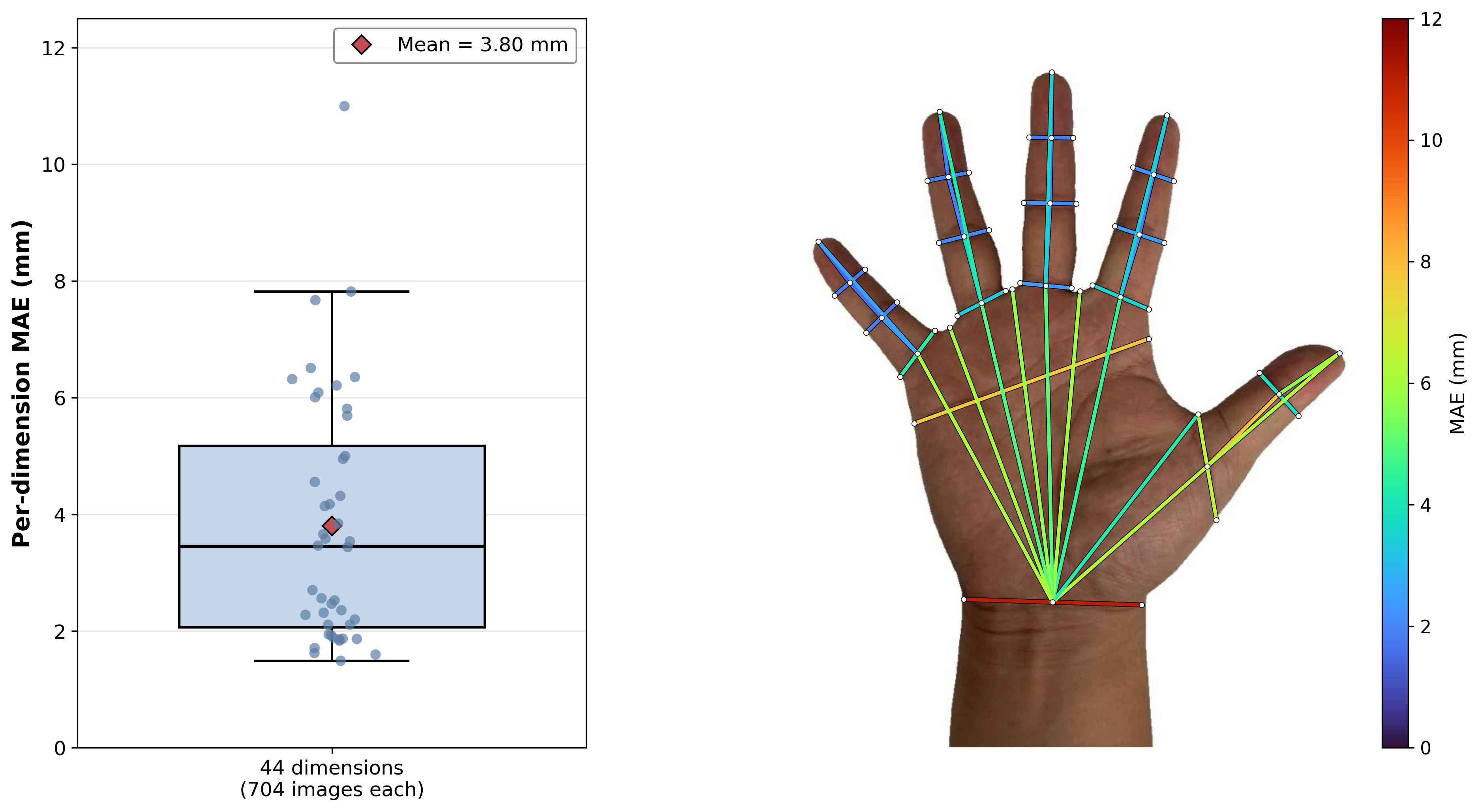}
\caption{Per-dimension MAE for $E_{16}$ across 44 dimensions (704 images, 45 participants). (a) Distribution (median 3.45\,mm, mean 3.80\,mm). (b) Spatial map in which each line connects a dimension's two endpoints and is colored by its MAE magnitude.}
\label{fig:e12_dim_summary}
\end{figure}

\begin{samepage}
\section{Per-Dimension Accuracy Breakdown}
\label{app:dim_breakdown}
\label{app:dim_full}

Table~\ref{tab:dim_full} defines all 44 dimensions and reports their absolute-error statistics on the same 704 complete captures; Fig.~\ref{fig:e12_dim_summary} shows their distribution and spatial pattern. Regional MAEs equally average the dimension-specific MAEs within D1--D5 (thumb), D6--D33 (non-thumb fingers), and D34--D44 (palm/wrist). The non-thumb group includes breadths, segment lengths, and complete digit lengths. Table~\ref{tab:mm_pp_summary} uses a separate length/width grouping.

\end{samepage}

\begin{table*}[!tp]
\centering
\footnotesize
\setlength{\tabcolsep}{3pt}
\begin{tabularx}{\textwidth}{@{}c>{\raggedright\arraybackslash}X l c c c c@{}}
\toprule
ID & Measurement & Endpoints & Mean $\pm$ SD & Median & $P_{95}$ & Range \\
\midrule
\multicolumn{7}{l}{\textit{Thumb (5)}} \\
D1 & Thumb IP breadth & $T_2\!\to T_3$ & $3.84\pm1.46$ & 3.77 & 6.57 & 0.98--10.01 \\
D2 & Thumb MCP breadth & $T_4\!\to T_5$ & $6.51\pm3.71$ & 6.45 & 12.63 & 0.06--18.17 \\
D3 & Thumb distal-segment length & $T_1\!\to\mathrm{MFK}_T$ & $5.69\pm3.09$ & 5.36 & 10.87 & 0.00--14.03 \\
D4 & Thumb proximal-segment length & $\mathrm{MFK}_T\!\to T_6$ & $7.82\pm4.43$ & 7.57 & 15.65 & 0.05--23.46 \\
D5 & Thumb total length & $T_1\!\to T_6$ & $6.32\pm4.08$ & 6.01 & 13.49 & 0.04--20.18 \\
\midrule
\multicolumn{7}{l}{\textit{Index finger (7)}} \\
D6 & Index DIP breadth & $I_2\!\to I_3$ & $2.28\pm1.00$ & 2.19 & 4.13 & 0.08--5.25 \\
D7 & Index PIP breadth & $I_4\!\to I_5$ & $2.53\pm0.99$ & 2.41 & 4.33 & 0.31--5.87 \\
D8 & Index MCP breadth & $I_6\!\to I_7$ & $3.58\pm1.62$ & 3.52 & 6.39 & 0.04--10.26 \\
D9 & Index distal-segment length & $I_1\!\to\mathrm{MFK}_I$ & $2.47\pm1.63$ & 2.30 & 5.31 & 0.01--7.76 \\
D10 & Index middle-segment length & $\mathrm{MFK}_I\!\to\mathrm{MSK}_I$ & $1.71\pm1.22$ & 1.57 & 3.79 & 0.00--5.81 \\
D11 & Index proximal-segment length & $\mathrm{MSK}_I\!\to\mathrm{MTK}_I$ & $2.11\pm1.54$ & 1.85 & 5.07 & 0.00--7.41 \\
D12 & Index total length & $I_1\!\to\mathrm{MTK}_I$ & $3.44\pm2.21$ & 3.31 & 7.14 & 0.03--11.81 \\
\midrule
\multicolumn{7}{l}{\textit{Middle finger (7)}} \\
D13 & Middle DIP breadth & $M_2\!\to M_3$ & $1.89\pm0.84$ & 1.82 & 3.31 & 0.02--6.28 \\
D14 & Middle PIP breadth & $M_4\!\to M_5$ & $2.20\pm0.88$ & 2.14 & 3.77 & 0.34--6.27 \\
D15 & Middle MCP breadth & $M_6\!\to M_7$ & $2.36\pm1.42$ & 2.26 & 4.96 & 0.01--7.19 \\
D16 & Middle distal-segment length & $M_1\!\to\mathrm{MFK}_M$ & $1.62\pm1.14$ & 1.43 & 3.58 & 0.00--5.55 \\
D17 & Middle middle-segment length & $\mathrm{MFK}_M\!\to\mathrm{MSK}_M$ & $1.84\pm1.38$ & 1.57 & 4.43 & 0.00--8.20 \\
D18 & Middle proximal-segment length & $\mathrm{MSK}_M\!\to\mathrm{MTK}_M$ & $2.70\pm1.74$ & 2.43 & 5.90 & 0.01--7.88 \\
D19 & Middle total length & $M_1\!\to\mathrm{MTK}_M$ & $3.54\pm2.29$ & 3.15 & 7.42 & 0.01--10.19 \\
\midrule
\multicolumn{7}{l}{\textit{Ring finger (7)}} \\
D20 & Ring DIP breadth & $R_2\!\to R_3$ & $1.93\pm0.73$ & 1.90 & 3.09 & 0.00--6.97 \\
D21 & Ring PIP breadth & $R_4\!\to R_5$ & $2.11\pm0.90$ & 2.02 & 3.75 & 0.01--7.14 \\
D22 & Ring MCP breadth & $R_6\!\to R_7$ & $3.46\pm1.85$ & 3.34 & 6.63 & 0.00--8.82 \\
D23 & Ring distal-segment length & $R_1\!\to\mathrm{MFK}_R$ & $1.87\pm1.50$ & 1.58 & 4.36 & 0.01--8.95 \\
D24 & Ring middle-segment length & $\mathrm{MFK}_R\!\to\mathrm{MSK}_R$ & $1.94\pm1.28$ & 1.76 & 4.36 & 0.01--6.01 \\
D25 & Ring proximal-segment length & $\mathrm{MSK}_R\!\to\mathrm{MTK}_R$ & $3.66\pm1.97$ & 3.68 & 6.33 & 0.03--12.41 \\
D26 & Ring total length & $R_1\!\to\mathrm{MTK}_R$ & $4.14\pm2.47$ & 4.03 & 8.57 & 0.01--11.20 \\
\midrule
\multicolumn{7}{l}{\textit{Little finger (7)}} \\
D27 & Little DIP breadth & $L_2\!\to L_3$ & $1.87\pm0.78$ & 1.84 & 3.21 & 0.04--4.79 \\
D28 & Little PIP breadth & $L_4\!\to L_5$ & $1.86\pm0.98$ & 1.80 & 3.63 & 0.01--7.88 \\
D29 & Little MCP breadth & $L_6\!\to L_7$ & $4.32\pm2.19$ & 4.26 & 8.19 & 0.05--12.75 \\
D30 & Little distal-segment length & $L_1\!\to\mathrm{MFK}_L$ & $1.49\pm1.30$ & 1.12 & 3.80 & 0.00--7.09 \\
D31 & Little middle-segment length & $\mathrm{MFK}_L\!\to\mathrm{MSK}_L$ & $1.60\pm1.12$ & 1.41 & 3.63 & 0.00--4.95 \\
D32 & Little proximal-segment length & $\mathrm{MSK}_L\!\to\mathrm{MTK}_L$ & $2.31\pm1.43$ & 2.23 & 5.05 & 0.01--7.09 \\
D33 & Little total length & $L_1\!\to\mathrm{MTK}_L$ & $2.56\pm2.05$ & 2.09 & 6.59 & 0.01--10.91 \\
\midrule
\multicolumn{7}{l}{\textit{Palm and wrist (11)}} \\
D34 & Metacarpal hand breadth & $P_1\!\to P_2$ & $7.67\pm2.42$ & 7.77 & 11.41 & 1.08--15.40 \\
D35 & Wrist breadth & $W_1\!\to W_2$ & $11.00\pm3.66$ & 10.57 & 17.95 & 2.79--25.64 \\
D36 & Thumb root--wrist-midpoint length & $T_6\!\to\mathrm{MoW}$ & $6.35\pm4.32$ & 5.95 & 14.88 & 0.01--20.12 \\
D37 & Thumb--index web--wrist-midpoint length & $T_5\!\to\mathrm{MoW}$ & $4.17\pm2.88$ & 3.68 & 9.55 & 0.02--12.88 \\
D38 & Index MCP--wrist-midpoint length & $\mathrm{MTK}_I\!\to\mathrm{MoW}$ & $4.56\pm3.68$ & 3.72 & 11.75 & 0.01--20.35 \\
D39 & Index--middle web--wrist-midpoint length & $\mathrm{RO}_1\!\to\mathrm{MoW}$ & $6.01\pm4.38$ & 5.13 & 14.68 & 0.00--18.86 \\
D40 & Middle MCP--wrist-midpoint length & $\mathrm{MTK}_M\!\to\mathrm{MoW}$ & $5.00\pm3.86$ & 4.13 & 12.94 & 0.01--16.52 \\
D41 & Middle--ring web--wrist-midpoint length & $\mathrm{RO}_2\!\to\mathrm{MoW}$ & $5.81\pm4.19$ & 5.00 & 14.48 & 0.03--18.62 \\
D42 & Ring MCP--wrist-midpoint length & $\mathrm{MTK}_R\!\to\mathrm{MoW}$ & $4.95\pm3.90$ & 4.12 & 13.29 & 0.00--19.33 \\
D43 & Ring--little web--wrist-midpoint length & $\mathrm{RO}_3\!\to\mathrm{MoW}$ & $6.09\pm4.29$ & 5.50 & 14.50 & 0.00--19.15 \\
D44 & Little MCP--wrist-midpoint length & $\mathrm{MTK}_L\!\to\mathrm{MoW}$ & $6.21\pm4.25$ & 5.42 & 14.48 & 0.05--20.16 \\
\bottomrule
\end{tabularx}
\caption{Definitions and per-dimension absolute-error statistics for $E_{16}$ on the 704 complete outputs. Endpoints use the notation in Tables~\ref{tab:keypoint_groups}--\ref{tab:derived_landmarks}; all statistics are in millimeters, and the Range column reports minimum and maximum errors.}
\label{tab:dim_full}
\end{table*}

\FloatBarrier
\section{Per-Keypoint Pixel Error by Anatomical Group}
\label{app:pixel_breakdown}
\begin{figure}[!htbp]
    \centering
    \subcaptionbox{A0\label{supp:vis:A0}}[0.48\columnwidth]{\includegraphics[width=0.34\columnwidth]{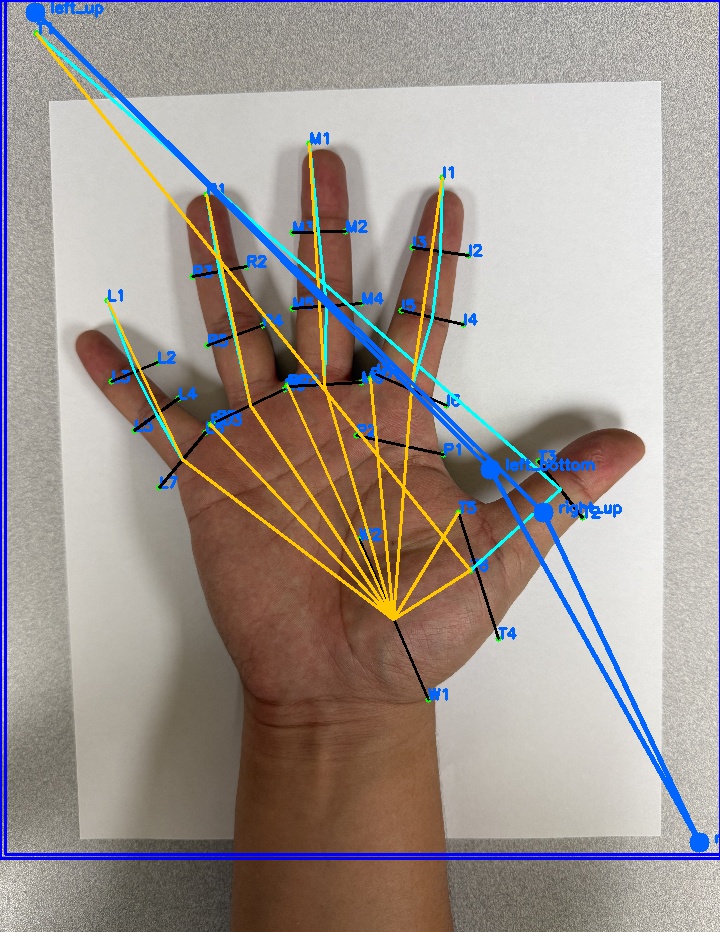}}\hfill
    \subcaptionbox{A1: +PR\label{supp:vis:A1}}[0.48\columnwidth]{\includegraphics[width=0.34\columnwidth]{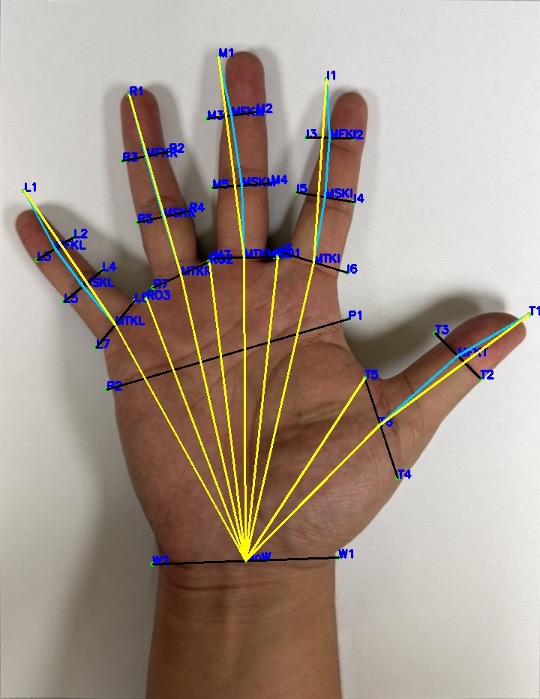}}
    \par\smallskip
    \subcaptionbox{A2: +BW\label{supp:vis:A2}}[0.48\columnwidth]{\includegraphics[width=0.34\columnwidth]{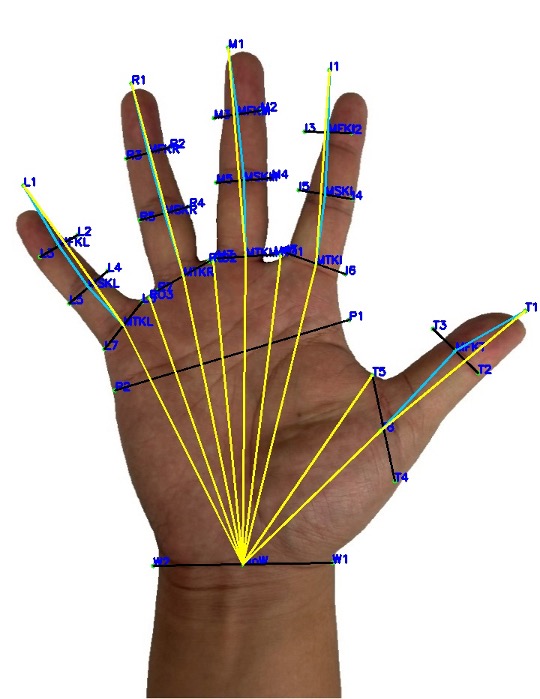}}\hfill
    \subcaptionbox{A3: +PP\label{supp:vis:A3}}[0.48\columnwidth]{\includegraphics[width=0.34\columnwidth]{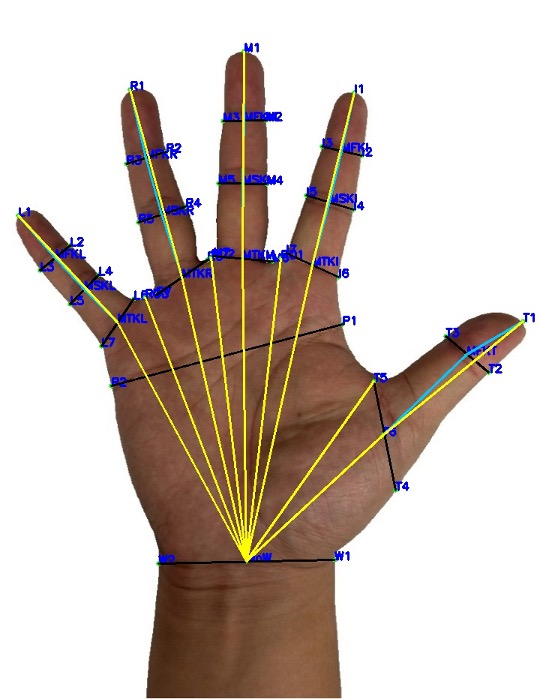}}
    \caption{Representative outputs across the progressive ablation configurations.}
    \label{fig:supp_ablation_outputs}
\end{figure}

Tables~\ref{tab:e15_vs_e16} and~\ref{tab:pixel_group_breakdown} compare pooled and regional pixel errors for $E_{15}$ (A2, without PP) and $E_{16}$ (A3, with PP) on 45 manually annotated images, one per participant (1{,}845 keypoint pairs). Mean error decreased on 44 of the 45 images. Figure~\ref{fig:supp_ablation_outputs} shows representative A0--A3 outputs.

\begin{table}[!htbp]
    \centering
    \footnotesize
    \setlength{\tabcolsep}{6pt}

    \begin{tabular}{lccc}
    \toprule
    Metric & $E_{15}$ (A2) & $E_{16}$ (A3) & $\%\Delta$ \\
    \midrule
    KP Mean (px)            & 11.18 & \textbf{7.83}  & $-29.94\%$ \\
    KP Std (px)             & 8.73  & \textbf{6.74}  & $-22.77\%$ \\
    Median error (px)       & 8.87  & \textbf{6.12}  & $-30.98\%$ \\
    IQR (px)                & 9.13  & \textbf{6.26}  & $-31.47\%$ \\
    $P(\text{error} \leq 5\,\text{px})$   & 21.52\% & \textbf{39.02\%} & $+81.3\%$ \\
    $P(\text{error} \leq 10\,\text{px})$  & 57.56\% & \textbf{75.77\%} & $+31.6\%$ \\
    \bottomrule
    \end{tabular}
    \caption{Pixel-level comparison of $E_{15}$ (A2, no PP) and $E_{16}$ (A3, with PP) on 45 annotated test images (1{,}845 keypoint pairs). $\%\Delta=(E_{16}-E_{15})/E_{15}$.}
    \label{tab:e15_vs_e16}
\end{table}

\begin{table}[!htbp]
    \centering
    \footnotesize
    \setlength{\tabcolsep}{5pt}

    \begin{tabular}{lccc}
    \toprule
    Group & $E_{15}$ (A2) & $E_{16}$ (A3) & $\%\Delta$ \\
    \midrule
    Index   & 8.57  & 5.59  & $-34.8\%$ \\
    Middle  & 8.70  & 6.05  & $-30.5\%$ \\
    Ring    & 10.81 & 6.90  & $-36.1\%$ \\
    Little  & 13.13 & 7.35  & $-44.0\%$ \\
    Thumb   & 17.35 & 14.54 & $-16.2\%$ \\
    Root    & 5.91  & 5.85  & $-1.0\%$ \\
    Palm    & 10.78 & 7.75  & $-28.2\%$ \\
    Wrist   & 13.28 & 9.83  & $-26.0\%$ \\
    \midrule
    All     & 11.18 & 7.83  & $-29.9\%$ \\
    \bottomrule
    \end{tabular}
    \caption{Mean keypoint pixel error (px) by anatomical group for $E_{15}$ (A2) and $E_{16}$ (A3) on 45 real test images.}
    \label{tab:pixel_group_breakdown}
\end{table}

\section{Paired Millimeter-Space PP Ablation}
\label{app:mm_pp_ablation}

\paragraph{Paired design.} $E_{15}$ (A2, no PP) and $E_{16}$ (A3, full PP; Appendix~\ref{app:pp_formulation}) use identical inputs, PR, BW, model weights, raw predictions, midpoint construction, dimension definitions, and metric conversion. Both completed the same 704 of 720 captures from all 45 participants; the remaining 16 failed in both configurations.

\paragraph{Aggregation and uncertainty.} Each dimension's MAE averages absolute errors against the two-operator mean caliper reference across the 704 captures. Overall and group MAEs equally weight their member dimensions (44 overall; 19 finger lengths, 14 finger widths, 11 palm/wrist). Captures have equal weight within dimensions, so participants with more completed captures contribute more observations. Define $\Delta=\mathrm{MAE}_{\mathrm{A3}}-\mathrm{MAE}_{\mathrm{A2}}$. To obtain paired participant-cluster bootstrap intervals, we resample 45 participants with replacement, retaining their common captures, dimensions, and both configurations together before recomputing MAE. We use 10{,}000 replicates, NumPy's \texttt{default\_rng} with seed 20260908, and the 2.5th/97.5th percentiles. Per-dimension intervals are exploratory without multiplicity adjustment. Tables~\ref{tab:mm_pp_summary} and~\ref{tab:mm_pp_dimensions} report all group and dimension results.

\begin{table}[!htbp]
\centering
\footnotesize
\setlength{\tabcolsep}{3pt}
\begin{tabularx}{\linewidth}{@{}>{\raggedright\arraybackslash}Xrrl@{}}
\toprule
Group ($n$) & A2 & A3 & $\Delta$ [95\% CI] \\
\midrule
Overall (44) & 3.707 & 3.804 & $+0.097\;[0.002, 0.193]$ \\
Finger length (19) & 3.196 & 3.097 & $-0.100\;[-0.235, 0.038]$ \\
Finger width (14) & 2.722 & 2.910 & $+0.188\;[0.035, 0.340]$ \\
Palm/wrist (11) & 5.842 & 6.165 & $+0.323\;[0.167, 0.494]$ \\
\bottomrule
\end{tabularx}
\caption{Paired PP ablation on 704 common completed captures from 45 participants. A2 and A3 columns report MAE; $n$ counts dimensions. All values are in millimeters, and negative $\Delta$ indicates improvement. Confidence intervals use participant-cluster resampling.}
\label{tab:mm_pp_summary}
\end{table}

\begin{table*}[!tp]
\centering
\footnotesize
\setlength{\tabcolsep}{4pt}
\begin{tabular}{@{}crrrcrrrr@{}}
\toprule
& \multicolumn{4}{c}{Mean absolute error} & \multicolumn{2}{c}{Signed bias} & \multicolumn{2}{c}{$P_{95}$ absolute error} \\
\cmidrule(lr){2-5}\cmidrule(lr){6-7}\cmidrule(l){8-9}
ID & A2 & A3 & $\Delta$ & 95\% CI for $\Delta$ & A2 & A3 & A2 & A3 \\
\midrule
D1 & 2.709 & 3.843 & $+1.134$ & $[0.775, 1.504]$ & $+2.700$ & $+3.809$ & 4.830 & 6.565 \\
D2 & 8.450 & 6.509 & $-1.941$ & $[-2.389, -1.501]$ & $+8.426$ & $+6.324$ & 15.552 & 12.629 \\
D3 & 3.280 & 5.691 & $+2.410$ & $[1.967, 2.878]$ & $-2.600$ & $-5.540$ & 7.641 & 10.866 \\
D4 & 11.083 & 7.819 & $-3.264$ & $[-4.055, -2.474]$ & $+10.884$ & $+7.318$ & 18.193 & 15.655 \\
D5 & 9.927 & 6.317 & $-3.610$ & $[-4.720, -2.540]$ & $+9.794$ & $+3.336$ & 17.745 & 13.486 \\
D6 & 2.323 & 2.278 & $-0.045$ & $[-0.310, 0.207]$ & $+2.313$ & $+2.278$ & 4.473 & 4.132 \\
D7 & 2.304 & 2.527 & $+0.223$ & $[-0.015, 0.468]$ & $+2.289$ & $+2.519$ & 4.236 & 4.334 \\
D8 & 3.758 & 3.585 & $-0.173$ & $[-0.503, 0.158]$ & $+3.758$ & $+3.557$ & 6.220 & 6.387 \\
D9 & 1.869 & 2.468 & $+0.599$ & $[0.407, 0.793]$ & $-1.419$ & $-2.191$ & 4.312 & 5.314 \\
D10 & 1.544 & 1.709 & $+0.166$ & $[0.023, 0.301]$ & $+0.911$ & $+0.655$ & 3.571 & 3.786 \\
D11 & 2.045 & 2.110 & $+0.064$ & $[-0.152, 0.291]$ & $+1.672$ & $+1.512$ & 4.693 & 5.070 \\
D12 & 2.547 & 3.442 & $+0.895$ & $[0.503, 1.275]$ & $+0.856$ & $-0.250$ & 5.872 & 7.138 \\
D13 & 1.755 & 1.888 & $+0.133$ & $[-0.095, 0.363]$ & $+1.732$ & $+1.886$ & 3.536 & 3.309 \\
D14 & 1.492 & 2.199 & $+0.708$ & $[0.458, 0.957]$ & $+1.446$ & $+2.199$ & 3.228 & 3.773 \\
D15 & 2.180 & 2.358 & $+0.178$ & $[-0.033, 0.384]$ & $+2.167$ & $+2.193$ & 4.264 & 4.965 \\
D16 & 1.464 & 1.625 & $+0.160$ & $[-0.010, 0.328]$ & $-0.410$ & $-1.040$ & 3.604 & 3.581 \\
D17 & 1.908 & 1.836 & $-0.072$ & $[-0.170, 0.027]$ & $+1.445$ & $+1.363$ & 4.483 & 4.428 \\
D18 & 2.437 & 2.702 & $+0.265$ & $[0.120, 0.403]$ & $+2.114$ & $+2.327$ & 5.704 & 5.905 \\
D19 & 3.768 & 3.539 & $-0.230$ & $[-0.542, 0.062]$ & $+3.198$ & $+2.705$ & 8.331 & 7.420 \\
D20 & 1.917 & 1.933 & $+0.015$ & $[-0.198, 0.227]$ & $+1.917$ & $+1.933$ & 3.563 & 3.094 \\
D21 & 1.497 & 2.107 & $+0.610$ & $[0.391, 0.823]$ & $+1.467$ & $+2.086$ & 3.069 & 3.747 \\
D22 & 3.004 & 3.465 & $+0.461$ & $[0.152, 0.747]$ & $+2.842$ & $+3.313$ & 5.899 & 6.634 \\
D23 & 1.784 & 1.865 & $+0.081$ & $[-0.098, 0.276]$ & $-0.255$ & $-0.671$ & 4.617 & 4.360 \\
D24 & 1.807 & 1.943 & $+0.136$ & $[-0.022, 0.295]$ & $+1.056$ & $+1.271$ & 4.138 & 4.357 \\
D25 & 3.140 & 3.662 & $+0.522$ & $[0.292, 0.747]$ & $+2.918$ & $+3.550$ & 5.949 & 6.329 \\
D26 & 3.779 & 4.141 & $+0.362$ & $[-0.157, 0.871]$ & $+3.527$ & $+3.957$ & 7.638 & 8.566 \\
D27 & 2.078 & 1.870 & $-0.208$ & $[-0.418, 0.005]$ & $+2.078$ & $+1.862$ & 3.928 & 3.207 \\
D28 & 1.781 & 1.858 & $+0.078$ & $[-0.156, 0.309]$ & $+1.722$ & $+1.816$ & 3.332 & 3.629 \\
D29 & 2.864 & 4.318 & $+1.454$ & $[1.091, 1.830]$ & $+2.851$ & $+4.286$ & 5.075 & 8.186 \\
D30 & 1.411 & 1.491 & $+0.080$ & $[-0.198, 0.378]$ & $-0.156$ & $-0.527$ & 3.775 & 3.804 \\
D31 & 1.690 & 1.598 & $-0.092$ & $[-0.303, 0.124]$ & $+1.317$ & $+1.322$ & 3.935 & 3.633 \\
D32 & 2.246 & 2.314 & $+0.067$ & $[-0.251, 0.391]$ & $+1.506$ & $+1.765$ & 4.408 & 5.053 \\
D33 & 3.001 & 2.562 & $-0.439$ & $[-1.147, 0.287]$ & $+2.012$ & $+2.067$ & 7.080 & 6.592 \\
D34 & 9.579 & 7.672 & $-1.907$ & $[-2.535, -1.248]$ & $+9.575$ & $+7.672$ & 14.912 & 11.412 \\
D35 & 8.267 & 10.996 & $+2.730$ & $[1.820, 3.687]$ & $+8.264$ & $+10.996$ & 12.708 & 17.947 \\
D36 & 5.387 & 6.353 & $+0.966$ & $[0.613, 1.328]$ & $-4.250$ & $-5.412$ & 12.416 & 14.878 \\
D37 & 4.010 & 4.174 & $+0.164$ & $[-0.469, 0.811]$ & $+2.121$ & $-0.394$ & 9.592 & 9.548 \\
D38 & 4.626 & 4.556 & $-0.070$ & $[-0.166, 0.024]$ & $+3.687$ & $+3.519$ & 12.172 & 11.746 \\
D39 & 5.956 & 6.007 & $+0.051$ & $[-0.061, 0.162]$ & $+5.054$ & $+5.077$ & 14.143 & 14.680 \\
D40 & 5.070 & 4.999 & $-0.071$ & $[-0.109, -0.033]$ & $+4.233$ & $+4.182$ & 13.055 & 12.942 \\
D41 & 5.575 & 5.810 & $+0.236$ & $[0.126, 0.349]$ & $+4.873$ & $+5.303$ & 14.199 & 14.479 \\
D42 & 4.931 & 4.952 & $+0.021$ & $[-0.071, 0.117]$ & $+4.060$ & $+4.173$ & 13.407 & 13.288 \\
D43 & 5.358 & 6.087 & $+0.729$ & $[0.496, 0.963]$ & $+4.445$ & $+5.339$ & 13.464 & 14.501 \\
D44 & 5.505 & 6.210 & $+0.705$ & $[0.496, 0.913]$ & $+5.036$ & $+5.956$ & 13.247 & 14.478 \\
\bottomrule
\end{tabular}
\caption{Complete dimension-level PP comparison on the same 704 captures. IDs follow Table~\ref{tab:dim_full}, and all values are in millimeters. $\Delta=\mathrm{MAE}_{\mathrm{A3}}-\mathrm{MAE}_{\mathrm{A2}}$; negative values indicate improvement. The paired participant-bootstrap intervals are exploratory, without multiplicity adjustment. Signed bias is the mean signed error (prediction minus caliper reference); $P_{95}$ is the per-dimension 95th percentile of per-capture absolute errors, using linear interpolation.}
\label{tab:mm_pp_dimensions}
\end{table*}

\FloatBarrier
\section{Full Ablation Configurations}
\label{app:ablation_deltas}

Table~\ref{tab:ablation_all} reports training and validation pose loss for all 16 configurations.

\begin{table}[!htbp]
    \centering
    \footnotesize
    \setlength{\tabcolsep}{4pt}

    \begin{tabularx}{\linewidth}{
    l c c c
    >{\centering\arraybackslash}X
    >{\centering\arraybackslash}X
    }

    \toprule
    ID & YOLO & Train aug. & Case &
    Pose (Train) &
    Pose (Val) \\
    \midrule
    \multicolumn{6}{c}{\textbf{Without training data augmentation}} \\
    \midrule
    E1 & v8 & no & A0 & 7.29355 & 3.42462 \\
    E2 & v8 & no & A1 & 0.94881 & 0.78883 \\
    E3 & v8 & no & A2 & 0.93568 & 0.83675 \\
    E4 & v8 & no & A3 & -- & -- \\
    E5 & v11 & no & A0 & 9.02002 & 8.49234 \\
    E6 & v11 & no & A1 & 1.55365 & 0.86025 \\
    E7 & v11 & no & A2 & 0.85474 & 0.79316 \\
    E8 & v11 & no & A3 & -- & -- \\
    \midrule
    \multicolumn{6}{c}{\textbf{With training data augmentation}} \\
    \midrule
    E9 & v8 & yes & A0 & 8.39867 & 10.78510 \\
    E10 & v8 & yes & A1 & 0.65540 & 0.62843 \\
    E11 & v8 & yes & A2 & 0.49995 & 0.63917 \\
    E12 & v8 & yes & A3 & -- & -- \\
    E13 & v11 & yes & A0 & 8.35649 & 6.40620 \\
    E14 & v11 & yes & A1 & 0.51637 & 0.70538 \\
    E15 & v11 & yes & A2 & 0.64615 & \textbf{0.63816} \\
    E16 & v11 & yes & A3 & -- & -- \\
    \bottomrule
    \end{tabularx}
    \parbox{\linewidth}{\footnotesize \textbf{Note:} A3 shares the A2 network and therefore has identical pose loss (``--'').}
    \caption{Training and validation pose loss for all 16 configurations.}
    \label{tab:ablation_all}
\end{table}

\section{Capture-Condition Sensitivity and Within-Person Variability}
\label{app:robustness_table}

Table~\ref{tab:robustness} reports descriptive MAEs and image counts by capture factor; these are not equivalence tests. Nominal illumination settings were pooled because captured brightness differed little.

\begin{table}[!htbp]
    \centering
    \footnotesize
    \setlength{\tabcolsep}{2.5pt}

    \begin{tabularx}{\columnwidth}{@{}l>{\raggedright\arraybackslash}Xccc@{}}
    \toprule
    Factor & $A$ / $B$ & \makecell{$N_A$ /\\$N_B$} & \makecell{MAE$_A$ /\\MAE$_B$} & $\Delta$ \\
    \midrule
    Device & Android / iPhone & 350 / 354 & 3.83 / 3.78 & $+0.05$ \\
    Background & Complex / Simple & 354 / 350 & 3.80 / 3.80 & $-0.00$ \\
    Angle & Top-down / Oblique & 346 / 358 & 4.18 / 3.44 & $+0.74$ \\
    \bottomrule
    \end{tabularx}
    \caption{Descriptive accuracy by capture factor among complete outputs. $N_A$ and $N_B$ count images, not independent participants; $\Delta$MAE $=$ MAE$_A-$MAE$_B$, and all MAE values are in millimeters.}
    \label{tab:robustness}
\end{table}

\subsection{Within-Person Capture Variability}
\label{app:within_person}

Among 45 participants, 38 contributed 16 completed captures and seven contributed 12--15. Within each participant and dimension, we computed the SD and coefficient of variation (CV; SD/mean prediction, expressed as a percentage). Across the 1,980 participant--dimension combinations, median/$P_{95}$ SD was 1.25/3.99\,mm and median/$P_{95}$ CV was 3.41/8.12\%. These repeated summaries are conditional on completion and describe capture variability, not caliper accuracy.

For each participant and dimension, we also computed the absolute difference between mean predictions at the two levels of each factor, then averaged across participants and equally across dimensions: 2.36\,mm for angle, 0.46\,mm for smartphone, and 0.88\,mm for background. Table~\ref{tab:within_person} gives per-dimension results; Appendix~\ref{app:angle_breakdown} gives caliper-referenced angle contrasts.

\begin{table*}[!tp]
\centering
\footnotesize
\setlength{\tabcolsep}{4pt}
\begin{tabularx}{\textwidth}{@{}c>{\raggedright\arraybackslash}Xrrrrr@{}}
\toprule
& & \multicolumn{2}{c}{Within-person CV (\%)} & \multicolumn{3}{c}{Mean absolute condition-mean difference (mm)} \\
\cmidrule(lr){3-4}\cmidrule(l){5-7}
ID & Measurement & Median & $P_{95}$ & Angle & Smartphone & Background \\
\midrule
D1 & Thumb IP breadth & 3.16 & 7.15 & 1.02 & 0.37 & 0.62 \\
D2 & Thumb MCP breadth & 4.39 & 8.42 & 1.88 & 0.84 & 1.23 \\
D3 & Thumb distal-segment length & 9.43 & 12.36 & 3.87 & 0.76 & 1.42 \\
D4 & Thumb proximal-segment length & 7.14 & 9.80 & 3.77 & 0.85 & 1.52 \\
D5 & Thumb total length & 7.93 & 10.17 & 7.73 & 1.28 & 2.24 \\
D6 & Index DIP breadth & 3.91 & 5.42 & 1.06 & 0.31 & 0.35 \\
D7 & Index PIP breadth & 3.09 & 4.51 & 0.81 & 0.38 & 0.42 \\
D8 & Index MCP breadth & 4.21 & 6.27 & 0.89 & 0.50 & 0.59 \\
D9 & Index distal-segment length & 5.50 & 7.85 & 2.27 & 0.30 & 0.54 \\
D10 & Index middle-segment length & 6.54 & 8.32 & 2.38 & 0.26 & 0.50 \\
D11 & Index proximal-segment length & 4.70 & 5.90 & 1.94 & 0.30 & 0.56 \\
D12 & Index total length & 5.27 & 6.61 & 6.50 & 0.66 & 1.08 \\
D13 & Middle DIP breadth & 3.11 & 4.72 & 0.84 & 0.23 & 0.26 \\
D14 & Middle PIP breadth & 3.13 & 4.78 & 0.95 & 0.27 & 0.32 \\
D15 & Middle MCP breadth & 5.21 & 8.67 & 1.17 & 0.44 & 0.55 \\
D16 & Middle distal-segment length & 3.66 & 5.27 & 1.37 & 0.30 & 0.53 \\
D17 & Middle middle-segment length & 5.18 & 7.43 & 2.27 & 0.28 & 0.53 \\
D18 & Middle proximal-segment length & 3.13 & 4.75 & 1.19 & 0.27 & 0.58 \\
D19 & Middle total length & 3.55 & 4.59 & 4.81 & 0.59 & 0.90 \\
D20 & Ring DIP breadth & 3.01 & 5.43 & 0.74 & 0.18 & 0.26 \\
D21 & Ring PIP breadth & 2.85 & 5.52 & 0.78 & 0.26 & 0.27 \\
D22 & Ring MCP breadth & 4.58 & 8.14 & 0.78 & 0.43 & 0.70 \\
D23 & Ring distal-segment length & 3.49 & 6.47 & 1.18 & 0.25 & 0.77 \\
D24 & Ring middle-segment length & 4.57 & 7.67 & 1.75 & 0.30 & 0.69 \\
D25 & Ring proximal-segment length & 2.62 & 5.33 & 0.65 & 0.26 & 0.67 \\
D26 & Ring total length & 2.61 & 3.93 & 3.41 & 0.58 & 0.87 \\
D27 & Little DIP breadth & 3.10 & 5.61 & 0.62 & 0.15 & 0.17 \\
D28 & Little PIP breadth & 3.15 & 8.14 & 0.72 & 0.26 & 0.31 \\
D29 & Little MCP breadth & 5.96 & 8.73 & 1.37 & 0.62 & 0.79 \\
D30 & Little distal-segment length & 3.24 & 6.93 & 0.89 & 0.25 & 0.73 \\
D31 & Little middle-segment length & 3.00 & 5.36 & 0.44 & 0.20 & 0.46 \\
D32 & Little proximal-segment length & 4.83 & 7.49 & 0.88 & 0.35 & 0.85 \\
D33 & Little total length & 2.49 & 4.45 & 1.96 & 0.60 & 1.13 \\
D34 & Metacarpal hand breadth & 1.48 & 2.51 & 1.57 & 0.46 & 0.66 \\
D35 & Wrist breadth & 2.97 & 5.00 & 2.73 & 0.57 & 1.68 \\
D36 & Thumb root--wrist-midpoint length & 2.35 & 4.88 & 1.70 & 0.67 & 1.30 \\
D37 & Thumb--index web--wrist-midpoint length & 2.01 & 4.32 & 1.59 & 0.64 & 1.35 \\
D38 & Index MCP--wrist-midpoint length & 2.40 & 3.77 & 4.68 & 0.58 & 1.40 \\
D39 & Index--middle web--wrist-midpoint length & 2.82 & 4.04 & 5.61 & 0.61 & 1.48 \\
D40 & Middle MCP--wrist-midpoint length & 2.28 & 3.87 & 4.66 & 0.54 & 1.41 \\
D41 & Middle--ring web--wrist-midpoint length & 2.31 & 3.79 & 4.66 & 0.59 & 1.46 \\
D42 & Ring MCP--wrist-midpoint length & 2.50 & 3.96 & 4.93 & 0.50 & 1.47 \\
D43 & Ring--little web--wrist-midpoint length & 2.41 & 4.84 & 4.64 & 0.56 & 1.51 \\
D44 & Little MCP--wrist-midpoint length & 2.34 & 4.27 & 4.27 & 0.53 & 1.54 \\
\bottomrule
\end{tabularx}
\caption{Within-person capture variability across 44 dimensions. For each dimension, median and $P_{95}$ summarize the CVs of 45 participants; each condition-mean difference is the average across participants of the absolute difference between their two condition-specific mean predictions. Summaries use the 704 completed captures. Dimension IDs and endpoints follow Table~\ref{tab:dim_full}.}
\label{tab:within_person}
\end{table*}

\FloatBarrier
\section{Dimension-Specific Capture-Angle Patterns}
\label{app:angle_breakdown}

For the 704 complete outputs (346 top-down; 358 oblique), let $\Delta_d=\mathrm{MAE}_{d,\mathrm{top\mbox{-}down}}-\mathrm{MAE}_{d,\mathrm{oblique}}$. The mean across dimensions was $+0.74$\,mm; 26 of 44 contrasts were positive and 18 were negative. Table~\ref{tab:angle_patterns} summarizes the largest and recurrent patterns.

\begin{table}[!htbp]
\centering
\footnotesize
\setlength{\tabcolsep}{3pt}
\begin{tabularx}{\columnwidth}{@{}>{\raggedright\arraybackslash}p{0.18\columnwidth}>{\raggedright\arraybackslash}X>{\raggedleft\arraybackslash}p{0.20\columnwidth}@{}}
\toprule
Pattern & Dimensions & $\Delta_d$ (mm) \\
\midrule
Top-down worse & Wrist-referenced dimensions (D38--D44) & $+2.92$ to $+4.28$ \\
& Middle/ring total lengths (D19, D26) & $+3.19$, $+2.97$ \\
& Thumb proximal/total lengths (D4, D5) & $+2.89$, $+2.22$ \\
\midrule
Oblique worse & Thumb distal length (D3) & $-3.74$ \\
& Index/middle/ring DIP and PIP breadths (D6, D7, D13, D14, D20, D21) & $-1.09$ to $-0.58$ \\
\midrule
Near zero & 12 of 44 dimensions & $|\Delta_d|\leq 0.50^{\dagger}$ \\
\bottomrule
\end{tabularx}
\parbox{\columnwidth}{\footnotesize $^{\dagger}$At the two-decimal precision shown in the current analysis; this descriptive cutoff is not an equivalence margin.}
\caption{Largest and recurrent dimension-specific capture-angle contrasts. Positive $\Delta_d$ indicates higher top-down MAE; all contrasts are descriptive. Dimension definitions appear in Table~\ref{tab:dim_full}.}
\label{tab:angle_patterns}
\end{table}


\section{Runtime Analysis}
\label{app:runtime}

Table~\ref{tab:raw_processing_time} reports $E_{16}$/A3 runtime on an Apple M3 (8-core CPU: 4 performance and 4 efficiency cores; 16 successes from 20 inputs) and an NVIDIA RTX 4000 Ada Generation GPU (20\,GB GDDR6; 704 complete test outputs). The workloads differ and do not constitute a matched hardware comparison. The BW runtime includes only cached-mask composition, excluding mask I/O.

\begin{table}[!htbp]
    \centering
    \footnotesize
    \setlength{\tabcolsep}{3.5pt}

    \begin{tabular}{@{}lcc@{}}
    \toprule
    Stage & CPU (M3) & GPU (RTX 4000 Ada) \\
    \midrule
    Total execution time      & $26.36 \pm 5.59$ & $3.86 \pm 0.61$ \\
    Perspective rectification & $24.98 \pm 5.25$ & $2.30 \pm 0.38$ \\
    Background whitening      & $\phantom{0}0.02 \pm 0.01$ & $0.02 \pm 0.01$ \\
    Inference and PP          & $\phantom{0}1.16 \pm 0.37$ & $1.51 \pm 0.31$ \\
    \bottomrule
    \end{tabular}
    \caption{Per-image stage runtime (mean $\pm$ SD, seconds). CPU: Apple M3, 16 successful images from a 20-image batch; GPU: RTX 4000 Ada, 704 complete outputs. BW reports mask-to-white composition only; cached-mask I/O is excluded.}
    \label{tab:raw_processing_time}
\end{table}

\section{Annotation-to-Caliper Discrepancy Analysis}
\label{app:oracle_analysis}

On one manually annotated rectified image from each of the 45 test participants, we derive all 44 dimensions using HandAnthro's midpoint construction and metric conversion. Let $y^{\mathrm{ann}}_{i,d}$ be the annotation-derived dimension and $\bar y^{\mathrm{cal}}_{i,d}$ the two-operator mean caliper reference. The pooled discrepancy is
\begin{equation}
\mathrm{MAE}_{\mathrm{annotation}}
=\frac{1}{45\times44}\sum_{i=1}^{45}\sum_{d=1}^{44}
\left|y^{\mathrm{ann}}_{i,d}-\bar y^{\mathrm{cal}}_{i,d}\right|.
\end{equation}
Across 1,980 errors, the mean was 3.38\,mm and the sample SD was 3.50\,mm. This diagnostic combines endpoint-definition, projection, annotation, rectification, and caliper discrepancies without isolating their contributions. Manual image annotations are not error-free, so this result is not an irreducible lower bound on automated measurement error.

\section{Inter-Operator Reliability of Manual Caliper Measurements}
\label{app:manual_reliability}

\begin{figure}[!htbp]
\centering
\includegraphics[width=\columnwidth]{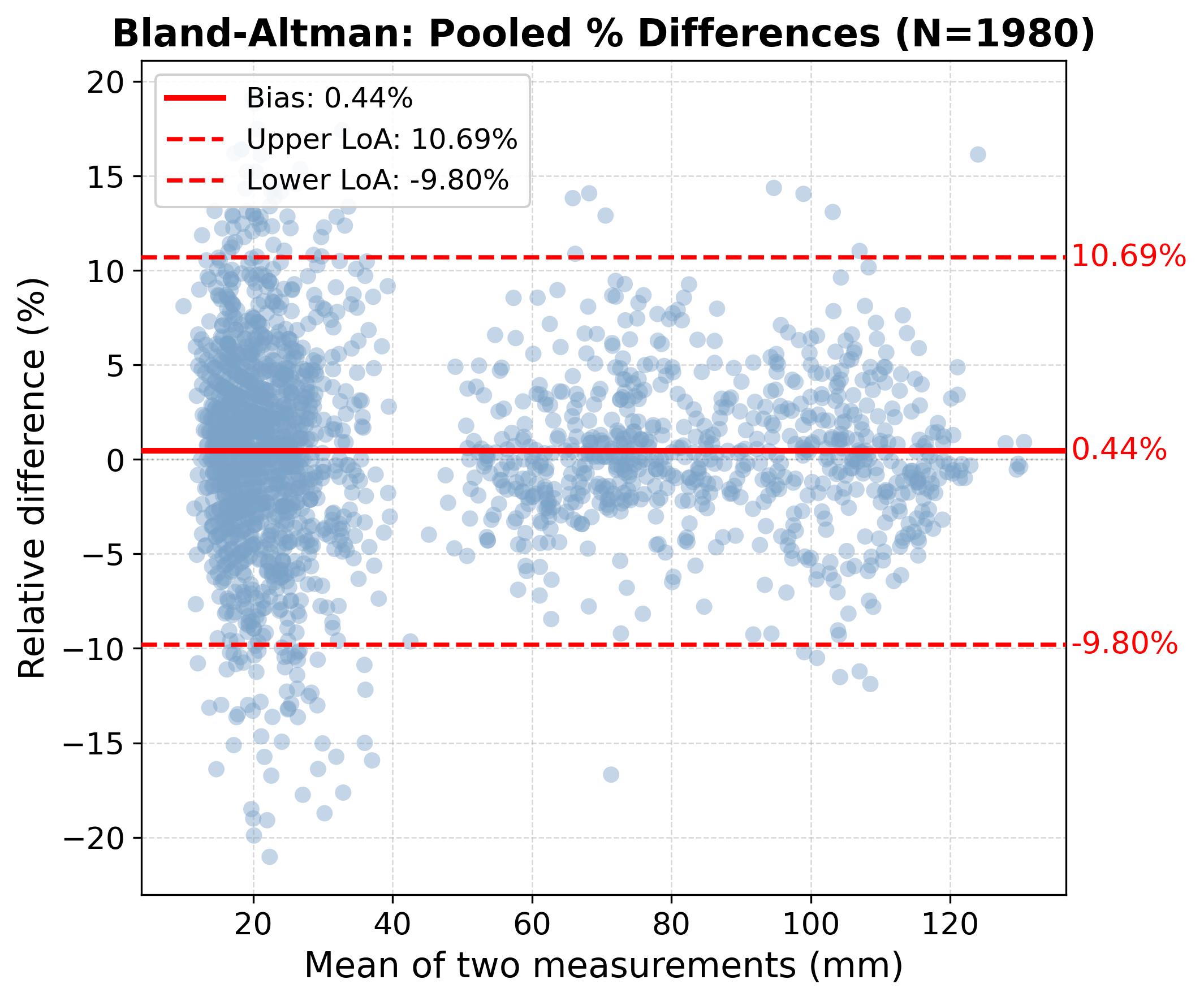}
\caption{Bland--Altman plot of signed relative differences between the two trained operators ($N=1{,}980$ across 45 hands and 44 dimensions). Horizontal lines indicate the pooled mean bias and 95\% limits of agreement.}
\label{Fig.Bland Altman plot}
\end{figure}

An operator estimated that measuring all 44 dimensions of one hand would take approximately 15 minutes. Two trained operators independently measured each of the 45 hands once for all 44 dimensions, yielding $N=1{,}980$ paired observations. For readings $x^{(1)}_{ij},x^{(2)}_{ij}$, define the pair mean $m_{ij}=(x^{(1)}_{ij}+x^{(2)}_{ij})/2$ and signed difference $d_{ij}=x^{(1)}_{ij}-x^{(2)}_{ij}$. The absolute difference is $|d_{ij}|$ and the absolute percent difference (APD) is $100|d_{ij}|/m_{ij}$. The Bland--Altman plot (Fig.~\ref{Fig.Bland Altman plot}) uses signed relative differences $r_{ij}=100d_{ij}/m_{ij}$ against $m_{ij}$, with pooled mean bias $\bar r$ and 95\% limits of agreement $\bar r\pm1.96s_r$, where $s_r$ is their sample SD.

Mean relative bias was $0.44\%$, with 95\% limits $[-9.80\%,+10.69\%]$. Median absolute difference was 0.90\,mm (IQR 1.50\,mm; $P_{95}=5.21$\,mm); median APD was 2.86\% (IQR 4.18\%; $P_{95}=11.00\%$).

\section{Unpaired Population-Level Comparison with the National Firefighter Reference}
\label{app:hsiao}

We compare sex-specific field-cohort summaries with Hsiao et al.'s independent national firefighter reference~\cite{hsiao2015firefighter_hand_Dim_Schema_design}. This unpaired comparison evaluates group-level plausibility, not individual accuracy or agreement.

\paragraph{Data and review.} The field cohort included 268 firefighters with one retained dominant-hand palmar image and a demographic record linked a priori (204 male, 64 female). Unlinked captures were excluded before processing, independently of algorithm outcome. Staff assisted with placement, app operation, and photography; unsuitable images were retaken, but attempt counts were not recorded. Retrospective desktop processing returned all 44 dimensions for 260 cases (97.0\%; 197 male, 63 female). Visual review of landmark overlays did not change coordinates, trigger recomputation, or exclude completed outputs. Hsiao et al.'s Table~2 reports means and SDs for 14 dimensions from 855 men and 88 women, measured on right-hand 2D flatbed scans with fingers open.

\paragraph{Dimension mapping.} Table~\ref{tab:hsiao_mapping} maps the reference dimensions to HandAnthro. Hand length sums palm and middle-finger lengths; non-thumb breadths use PIP landmarks. Metacarpal hand breadth is an explicit proxy for palm breadth because the pipeline has no diagonal palm-crease breadth.

\begin{table}[!htbp]
\centering
\footnotesize
\setlength{\tabcolsep}{4pt}
\begin{tabularx}{\linewidth}{l l X}
\toprule
Reference dimension & HandAnthro & Note \\
\midrule
Hand length & D40\,+\,D19 & palm length + middle-finger length \\
Hand breadth & D34 & across the metacarpals \\
Palm length & D40 & wrist-midpoint to middle MCP \\
Palm breadth & D34 & proxy (no palm-crease breadth) \\
Thumb length & D5 & tip to thumb root \\
Thumb breadth & D1 & IP-joint breadth \\
Index length & D12 & tip to MCP base \\
Index breadth & D7 & PIP-joint breadth \\
Middle length & D19 & tip to MCP base \\
Middle breadth & D14 & PIP-joint breadth \\
Ring length & D26 & tip to MCP base \\
Ring breadth & D21 & PIP-joint breadth \\
Pinky length & D33 & tip to MCP base \\
Pinky breadth & D28 & PIP-joint breadth \\
\bottomrule
\end{tabularx}
\caption{Mapping from Hsiao (2015) Table~2 dimensions to HandAnthro outputs.}
\label{tab:hsiao_mapping}
\end{table}

\paragraph{Method.} For each dimension and sex, Table~\ref{tab:hsiao_full} reports mean (SD), signed mean differences in millimeters and percent (HandAnthro minus reference), Hedges' $g$, and Welch two-sample test significance after Benjamini--Hochberg correction across all 28 contrasts. The descriptive count within 5\% uses unrounded relative differences.

\begin{table}[!htbp]
\centering
\footnotesize
\begin{minipage}[t]{\columnwidth}
\centering
\textbf{(a) Men}\par\smallskip
\setlength{\tabcolsep}{0.95pt}
\begin{tabular}{@{}lrrrrrc@{}}
\toprule
Dimension & Ours & Ref & $\Delta$ & $\Delta\%$ & $g$ & FDR \\
\midrule
Hand length    & 204.9 (10.6) & 197.6 (9.3) & $+7.33$ & $+3.7$ & $+0.77$ & yes \\
Hand breadth   & 100.8 (6.0) & 97.2 (4.6) & $+3.58$ & $+3.7$ & $+0.73$ & yes \\
Palm length    & 121.5 (6.0) & 113.8 (5.8) & $+7.75$ & $+6.8$ & $+1.33$ & yes \\
Palm breadth   & 100.8 (6.0) & 96.0 (4.6) & $+4.78$ & $+5.0$ & $+0.98$ & yes \\
Thumb length   & 76.0 (5.8) & 70.8 (4.3) & $+5.18$ & $+7.3$ & $+1.12$ & yes \\
Thumb breadth  & 27.7 (7.4) & 24.4 (1.6) & $+3.31$ & $+13.6$ & $+0.94$ & yes \\
Index length   & 75.8 (5.6) & 75.8 (4.4) & $+0.02$ & $+0.0$ & $+0.01$ & no \\
Index breadth  & 24.3 (1.8) & 22.7 (1.6) & $+1.63$ & $+7.2$ & $+0.99$ & yes \\
Middle length  & 83.4 (5.8) & 83.8 (4.6) & $-0.42$ & $-0.5$ & $-0.09$ & no \\
Middle breadth & 24.2 (1.9) & 22.4 (1.7) & $+1.82$ & $+8.1$ & $+1.05$ & yes \\
Ring length    & 77.0 (5.9) & 79.6 (4.5) & $-2.60$ & $-3.3$ & $-0.54$ & yes \\
Ring breadth   & 22.0 (1.6) & 21.7 (1.6) & $+0.30$ & $+1.4$ & $+0.19$ & yes \\
Pinky length   & 61.7 (5.6) & 65.2 (4.3) & $-3.49$ & $-5.4$ & $-0.76$ & yes \\
Pinky breadth  & 20.6 (1.7) & 19.8 (1.5) & $+0.79$ & $+4.0$ & $+0.51$ & yes \\
\bottomrule
\end{tabular}
\end{minipage}
\par\medskip
\begin{minipage}[t]{\columnwidth}
\centering
\textbf{(b) Women}\par\smallskip
\setlength{\tabcolsep}{0.95pt}
\begin{tabular}{@{}lrrrrrc@{}}
\toprule
Dimension & Ours & Ref & $\Delta$ & $\Delta\%$ & $g$ & FDR \\
\midrule
Hand length    & 184.4 (8.9) & 182.7 (8.7) & $+1.71$ & $+0.9$ & $+0.19$ & no \\
Hand breadth   & 88.1 (5.4) & 87.4 (4.2) & $+0.72$ & $+0.8$ & $+0.15$ & no \\
Palm length    & 108.0 (5.3) & 104.0 (5.7) & $+4.03$ & $+3.9$ & $+0.72$ & yes \\
Palm breadth   & 88.1 (5.4) & 85.3 (4.2) & $+2.82$ & $+3.3$ & $+0.59$ & yes \\
Thumb length   & 65.3 (5.2) & 64.8 (4.1) & $+0.49$ & $+0.8$ & $+0.11$ & no \\
Thumb breadth  & 22.1 (2.1) & 21.5 (1.5) & $+0.65$ & $+3.0$ & $+0.37$ & no \\
Index length   & 69.6 (4.8) & 71.3 (4.2) & $-1.68$ & $-2.4$ & $-0.37$ & yes \\
Index breadth  & 20.2 (1.4) & 20.5 (1.2) & $-0.26$ & $-1.3$ & $-0.20$ & no \\
Middle length  & 76.4 (4.5) & 78.6 (4.5) & $-2.21$ & $-2.8$ & $-0.49$ & yes \\
Middle breadth & 20.2 (1.2) & 20.3 (1.3) & $-0.14$ & $-0.7$ & $-0.11$ & no \\
Ring length    & 70.2 (5.0) & 74.0 (4.5) & $-3.77$ & $-5.1$ & $-0.80$ & yes \\
Ring breadth   & 18.9 (1.7) & 19.4 (1.2) & $-0.49$ & $-2.5$ & $-0.34$ & no \\
Pinky length   & 55.5 (5.9) & 60.4 (4.3) & $-4.88$ & $-8.1$ & $-0.97$ & yes \\
Pinky breadth  & 17.3 (1.6) & 17.5 (1.2) & $-0.24$ & $-1.4$ & $-0.17$ & no \\
\bottomrule
\end{tabular}
\end{minipage}
\caption{Sex-specific comparison of 260 successful dominant-hand palmar cases (197 male and 63 female) with Hsiao et al. Ours and Ref are mean (SD) in millimeters; $\Delta=$ Ours$-$Ref, $g$ is Hedges' $g$, and FDR indicates Benjamini--Hochberg significance at $q<0.05$.}
\label{tab:hsiao_full}
\end{table}

Across 28 contrasts, the mean absolute difference was 2.40\,mm (median 1.77\,mm), and 20/28 were within 5\% of the reference. Mean absolute Hedges' $g$ was 0.56; 18/28 contrasts remained significant after correction. The 5\% cutoff is descriptive, not an equivalence margin.

\paragraph{Interpretation limits.} The independent cohorts differ in recruitment, year, hand laterality (dominant versus right), and measurement protocol; palm breadth also uses a proxy. Their offsets cannot separate measurement effects from cohort differences. Reference summaries permit comparison of means and SDs only; statistical significance does not establish agreement.

\FloatBarrier
\section{Pipeline Failure-Mode Analysis}
\label{app:failure_modes}

\paragraph{Denominators.} PR and end-to-end completion were both 704/720 (97.8\%): every rectified capture returned all 44 dimensions. Accuracy uses these 704 captures, with no imputed errors for the 16 failures.

\paragraph{Controlled captures.} Visual review of all 16 PR failures found the wrist crease on or beyond a paper edge: 15 hands crossed a long edge and one sat low on the bottom edge. One or both lower paper prompts fell outside the sheet, the paper mask included background, and no valid quadrilateral was recovered. The complete sheet was visible in every failed frame. These failures involved seven participants (4, 4, 3, 2, 1, 1, and 1 captures); three of these participants had failures with both smartphones.

\paragraph{Field captures.} All eight incomplete field captures failed during PR. Post-hoc review found five hands placed too high or low, with adaptive prompts outside the sheet; one incompletely framed sheet; one hand extending beyond the paper; and one covered paper corner.

\section{Notes on the Literature Comparison}
\label{app:literature_comparison_notes}
These notes accompany Table~\ref{tab:literature_measurement_context} in the main paper.

Automation refers to dimension extraction after acquisition without per-hand manual landmark annotation or digitization. Equipment lists capture hardware; computing and software are additionally required.
Only values explicitly reported as hand-dimension MAE are reproduced. A dash indicates no such value is tabulated: Han and Park report dimension-specific MAD, and Nguyen et al. evaluate gauge-block calibration.
Han and Park claim 18 dimensions but tabulate 17.
HandAnthro MAEs use the same 704 complete captures from 45 participants. Regional groups comprise 28 non-thumb, five thumb, and 11 palm/wrist dimensions; each mean weights its member dimensions equally.